\RequirePackage{expl3}
\ExplSyntaxOn
\providecommand{\NewStructureName}[1]{}
\providecommand{\AssignStructureRole}[2]{}
\providecommand{\NewTaggingSocket}[2]{
  \int_if_exist:cF { c__tagsupport_ #1 _args_int }
    { \int_const:cn { c__tagsupport_ #1 _args_int } { #2 } }
}
\providecommand{\NewTaggingSocketPlug}[3]{}
\providecommand{\AssignTaggingSocketPlug}[2]{}
\providecommand{\UseTaggingSocket}[1]{}
\providecommand{\UseStructureName}[1]{}
\providecommand{\tagstructbegin}[1]{}

\ExplSyntaxOff
\documentclass[]{fairmeta}

\let\cite\citep

\usepackage{amsmath}
\usepackage{amssymb}
\usepackage{wrapfig}
\usepackage{enumitem}
\usepackage{pifont}
\usepackage{makecell}
\usepackage{threeparttable}
\usepackage{adjustbox}
\usepackage{colortbl}
\usepackage{pgfplots}
\pgfplotsset{compat=1.18}
\usepackage{float}

\graphicspath{{fig/}}

\newcommand{\filegram}{\textsc{FileGram}}
\newcommand{\filegramengine}{\textsc{FileGramEngine}}
\newcommand{\filegrambench}{\textsc{FileGramBench}}
\newcommand{\filegramqa}{\textsc{FileGramQA}}
\newcommand{\filegramos}{\textsc{FileGramOS}}

\definecolor{ccheck}{HTML}{1a8754}
\definecolor{ccross}{HTML}{c0392b}
\definecolor{cpartial}{HTML}{b8860b}
\definecolor{oursrow}{HTML}{fffde7}

\definecolor{cblue}{HTML}{1565c0}
\definecolor{cpurple}{HTML}{6a1b9a}
\definecolor{cred}{HTML}{c62828}

\newcommand{\cmark}{{\color{ccheck}\ding{51}}}
\newcommand{\xmark}{{\color{ccross}\ding{55}}}
\newcommand{\pmark}{{\color{cpartial}$\circ$}}

\newcommand{\profRecCount}{320}

\newcommand{\fpCount}{560}
\newcommand{\cfCount}{560}

\newcommand{\fileGrdCount}{550}
\newcommand{\visGrdCount}{100}

\title{FileGram: Grounding Agent Personalization in File-System Behavioral Traces}

\author[1,2]{Shuai Liu}
\author[1]{Shulin Tian}
\author[1,2]{Kairui Hu}
\author[1]{Yuhao Dong}
\author[1]{Zhe Yang}
\author[1]{Bo Li}
\author[1]{Jingkang Yang}
\author[1,2]{Chen Change Loy\textsuperscript{$\dagger$}}
\author[1]{Ziwei Liu\textsuperscript{$\dagger$}}

\affiliation[1]{S-Lab, Nanyang Technological University}
\affiliation[2]{Synvo AI}

\contribution[\dagger]{\ Corresponding authors}

\abstract{Coworking AI agents operating within local file systems are rapidly emerging as a paradigm in human-AI interaction. Since users exhibit highly diverse workflows, personalization is essential for tight collaboration and a seamless user experience. However, effective personalization is limited by severe data constraints, since strict privacy barriers and the inherent difficulty of jointly collecting multimodal real-world traces preclude the creation of scalable training data and comprehensive evaluation suites. Consequently, existing methods remain interaction-centric and overlook dense behavioral traces embedded in file-system operations. To bridge this gap, we propose \textbf{FileGram}, a comprehensive framework that grounds agent memory and personalization in file-system behavioral traces. FileGram comprises three core components to overcome current data and evaluation bottlenecks. 1) \textbf{FileGramEngine}, a scalable, persona-driven data engine that simulates realistic workflows to generate fine-grained, multimodal action sequences at scale. 2) \textbf{FileGramBench}, a diagnostic benchmark grounded in file-system behavioral traces. It evaluates memory systems across profile reconstruction, trace disentanglement, persona drift detection, and multimodal grounding. 3) \textbf{FileGramOS}, a bottom-up memory architecture that builds user profiles directly from atomic actions and content deltas rather than dialogue summaries. It encodes these traces into procedural, semantic, and episodic channels with query-time abstraction. Extensive experiments show that FileGramBench remains challenging for state-of-the-art memory systems. Our results also demonstrate the effectiveness of FileGramEngine and FileGramOS. By open-sourcing our framework, we aim to pave the way for future research on personalized memory-centric file-system agents.}

\date{April 6, 2026}
\metadata[Project Page]{\url{https://filegram.choiszt.com}}
\metadata[Code]{\url{https://github.com/Synvo-ai/FileGram}}

\begin{document}
\maketitle

\section{Introduction}
\label{sec:intro}
Driven by recent advancements in OS-level assistants, AI agents are rapidly evolving from conversational interfaces into integrated file-system coworkers. However, seamless human-AI collaboration requires transcending the execution of isolated commands. As users exhibit profound variability in their workflows, organizational habits, and execution styles, adapting to these distinct preferences is essential for agents to continuously align with long-term user behavior. Effective personalization thus requires grounding in two complementary signals from the file system \cite{lewis2020retrieval, park2023generative}: \emph{behavioral traces}, the sequence of operations a user performs such as reading, creating, and reorganizing files, and \emph{content deltas}, the incremental outputs a user actually produces and edits, which carry far stronger personal signatures than externally sourced materials like downloaded references and pre-existing templates. By inferring stable preferences from these file-level signals rather than transient dialogue, agents can achieve the reliable adaptation necessary for practical, everyday coworking.


Despite its importance, personalized behavioral adaptation is severely hindered by bottlenecks in data, evaluation, and methodology. First, regarding \emph{data}, collecting real-world, multimodal, and long-trajectory file-system data is prohibitively difficult \cite{xie2024osworld, mu2025gui360circ}. Strict privacy constraints and the absence of scalable collection strategies limit the capture of diverse user preferences. Second, for \emph{evaluation}, as shown in \cref{tab:benchmark-comparison}, existing benchmarks~\cite{wu2024longmemeval, hu2025memagentbench} heavily prioritize conversational recall or isolated GUI success rates \cite{zhou2023webarena, deng2023mind2web, Mialon2023GAIA}, overlooking memory-centric, personalized behavior understanding tasks. Finally, in terms of \emph{methodology}, mainstream memory architectures~\cite{chhikara2025mem0,rasmussen2025zep,li2025memos} remain fundamentally interaction-centric. By relying on top-down dialogue summaries, they lack the bottom-up architecture required to distill procedural behavior patterns from continuous file-system operations \cite{zeng2024agenttuning}. Document-centric methods~\cite{mathew2021docvqa,ma2024mmlongbenchdoc,han2025mdocagent} treat files as fixed knowledge artifacts, agnostic to \emph{who} produced them, and recent edit-based preference learning~\cite{gao2024aligningllmagentslearning} remains limited to single-turn generation. \filegram{} generalizes these insights to the file-system scale, jointly modeling atomic actions and content deltas across long-horizon, multimodal trajectories.

To address these bottlenecks, we propose \textbf{\filegram{}}, a unified framework designed to ground agent memory and personalization in file-system behavioral traces. This framework tackles the challenges through three core components. First, to overcome data scarcity, \textbf{\filegramengine{}} simulates multimodal file-system behavioral traces across realistic scenarios to enable scalable data generation. Second, for robust evaluation, \textbf{\filegrambench{}} serves as the first benchmark dedicated to memory-centric personalization tasks based on file-system operations. It provides four distinct evaluation tracks and 16 attributes spanning procedural, semantic, and episodic memory capabilities. Finally, to advance methodology, \textbf{\filegramos{}} introduces the first bottom-up architecture that constructs user profiles directly from atomic actions and content deltas rather than relying on top-down dialogue summaries. Together, these components establish a comprehensive foundation to evaluate and develop the next generation of memory-centric personalized AI coworkers.

Extensive experiments on \filegrambench{} reveal that existing memory systems struggle with file-system personalization: context-based and narrative-first baselines top out at 48--50\% accuracy, while multimodal methods fare even worse at 44.7\%. Our \filegramos{} achieves 59.6\% by preserving atomic actions and content deltas in a bottom-up architecture. Further analysis exposes a clear capability hierarchy, where current methods show partial competence in behavioral understanding but fail at shift attribution and multimodal grounding. Through \filegram{}, we aim to provide the essential data, evaluation, and structural foundation to drive the development of truly adaptive AI coworkers.

\begin{figure*}[t]
\centering
\includegraphics[width=\linewidth]{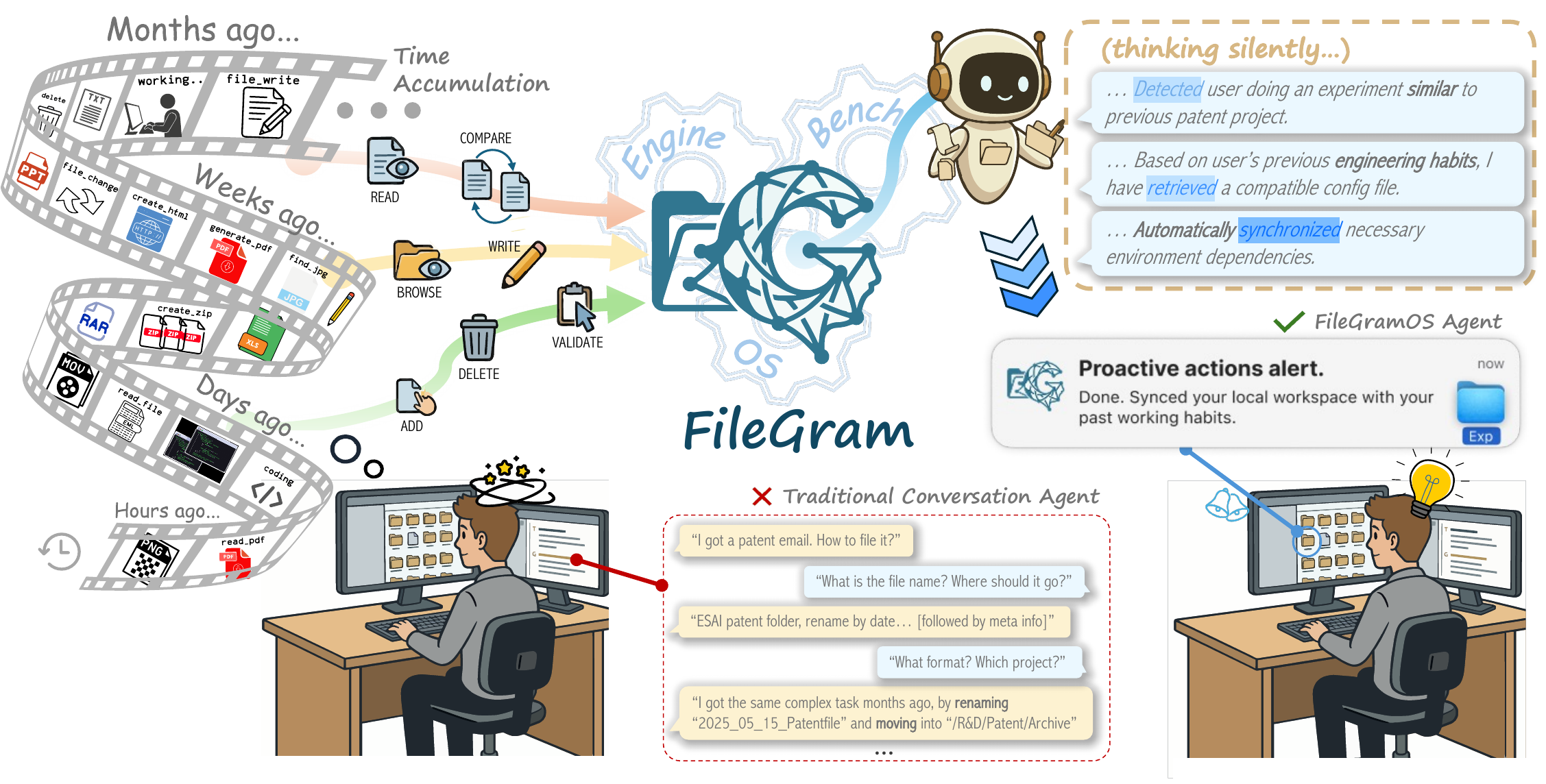}
\caption{\textbf{Overview of the FileGram Project.} FileGram introduces a personalized AI coworker natively integrated into the user file system. By consolidating cross-session activities and file outputs into long-term behavioral memory, the agent infers intent and proactively synchronizes workspaces, establishing a new paradigm for real-world interactive coworking.}
\label{fig:teaser}
\end{figure*}

\section{Related Work}
\label{sec:related}

\smallskip\noindent\textbf{Benchmarks for Agents and Memory.}
Prior benchmarks evaluate memory through two dominant paradigms: conversational recall and environmental task execution, as shown in \cref{tab:benchmark-comparison}. Conversational datasets focus heavily on static semantic retrieval over extended text dialogues~\cite{wu2024longmemeval,maharana2024locomo}, inherently stripping away the procedural context of real workflows. Conversely, execution-driven benchmarks situate agents in realistic operating systems or web interfaces but treat memory as a latent variable implicitly measured by objective task success~\cite{zhou2023webarena,xie2024osworld}. While recent trajectory-aware benchmarks~\cite{zhao2026amabench,he2026memoryarena} evaluate memory strictly for universal reasoning and generic fact retention, \filegrambench{} shifts this paradigm toward personalization. It provides the first controllable suite to evaluate how effectively agents infer and predict user-specific behaviors directly from longitudinal file-system traces.


\smallskip\noindent\textbf{Memory System and Personalization for Agents.}
Existing architectures predominantly extract explicit facts and relational structures from conversational histories~\cite{packer2023memgpt,chhikara2025mem0,rasmussen2025zep}, remaining fundamentally disconnected from the user's operational environment. While recent advancements in multimodal perception~\cite{lu2026mma,lin2025hippomm} and trajectory tracking~\cite{li2025memos,fang2025memp} capture temporal dynamics, they typically model these dimensions in isolation or within highly constrained, simulation-based environments such as online shopping or social media~\cite{wang2025customerr1, jin2025twice}. Crucially, no existing framework utilizes granular file-system activities to jointly sustain the procedural, semantic, and episodic memory required for continuous coworking adaptation. \filegramos{} bridges this gap by directly encoding atomic file-system actions and content deltas into a unified, three-channel memory framework for robust behavioral pattern extraction.

\begin{table}[!t]
\caption{Comparison of \filegrambench{} with representative benchmarks. Only \filegrambench{} jointly provides multimodal content, persistent memory, and file-system behavioral traces with controlled profiles. \textbf{Columns}: MS = multi-session; MM = multimodal; UP = user profile; Me = explicit memory component; FR = fact retrieval; Re = reasoning; KM = knowledge management; Pe = personalization.}
\label{tab:benchmark-comparison}
\centering
\renewcommand{\arraystretch}{1.05}
\setlength{\tabcolsep}{3.5pt}
{\footnotesize
\begin{adjustbox}{width=0.8\textwidth}
\begin{tabular}{@{}l c r ccc c cccc@{}}
\toprule
& & & \multicolumn{3}{c}{\textbf{Data}} & & \multicolumn{4}{c}{\textbf{Evaluation}} \\
\cmidrule(lr){4-6} \cmidrule(lr){8-11}
\textbf{Benchmark}
  & \textbf{Type}
  & \textbf{\#QA}
  & \textbf{MS}
  & \textbf{MM}
  & \textbf{UP}
  & \textbf{Me}
  & \textbf{FR}
  & \textbf{Re}
  & \textbf{KM}
  & \textbf{Pe} \\
\midrule
DuLeMon~\cite{xu2022dulemon}
  & Conv. & -- & \cmark & \xmark & \cmark & \cmark
  & \cmark & \xmark & \xmark & \cmark \\
DialogBench~\cite{ou2024dialogbench}
  & Conv. & 9.8K & \cmark & \xmark & \xmark & \cmark
  & \cmark & \xmark & \xmark & \xmark \\
MemoryBank~\cite{zhong2024memorybank}
  & Conv. & 194 & \cmark & \xmark & \cmark & \cmark
  & \cmark & \xmark & \cmark & \cmark \\
LongMemEval~\cite{wu2024longmemeval}
  & Conv. & 500 & \cmark & \xmark & \cmark & \cmark
  & \cmark & \cmark & \cmark & \xmark \\
MemAgentBench~\cite{hu2025memagentbench}
  & Conv. & 146 & \cmark & \xmark & \xmark & \cmark
  & \cmark & \xmark & \cmark & \xmark \\
MMDU~\cite{liu2024mmdu}
  & Conv. & 1.6K & \xmark & \cmark & \xmark & \cmark
  & \cmark & \xmark & \xmark & \xmark \\
LoCoMo~\cite{maharana2024locomo}
  & Conv. & 7.5K & \cmark & \cmark & \cmark & \cmark
  & \cmark & \cmark & \cmark & \cmark \\
MMRC~\cite{xue2025mmrc}
  & Conv. & 28.7K & \cmark & \cmark & \xmark & \cmark
  & \cmark & \cmark & \cmark & \xmark \\
Mem-Gallery~\cite{bei2026memgallery}
  & Conv. & 1.7K & \cmark & \cmark & \cmark & \cmark
  & \cmark & \cmark & \cmark & \xmark \\
\midrule
OSWorld~\cite{xie2024osworld}
  & GUI & 369 & \xmark & \cmark & \xmark & \xmark
  & \xmark & \xmark & \xmark & \xmark \\
OfficeBench~\cite{wang2024officebench}
  & GUI & 300 & \xmark & \cmark & \xmark & \xmark
  & \xmark & \xmark & \xmark & \xmark \\
MEMTRACK~\cite{deshpande2025memtrack}
  & Agent & 47 & \cmark & \xmark & \xmark & \cmark
  & \cmark & \cmark & \xmark & \xmark \\
AgencyBench~\cite{li2026agencybench}
  & Agent & 138 & \cmark & \xmark & \xmark & \cmark
  & \cmark & \xmark & \cmark & \xmark \\
Evo-Memory~\cite{wei2025evomemory}
  & Agent & -- & \cmark & \xmark & \xmark & \cmark
  & \cmark & \cmark & \cmark & \xmark \\
MemoryArena~\cite{he2026memoryarena}
  & Agent & 766 & \cmark & \xmark & \xmark & \cmark
  & \cmark & \cmark & \cmark & \xmark \\
\midrule
\rowcolor{oursrow}
\textbf{\filegrambench{}} {\scriptsize(\textbf{Ours})}
  & \textbf{File} & \textbf{4.6K} & \cmark & \cmark & \cmark & \cmark
  & \cmark & \cmark & \cmark & \cmark \\
\bottomrule
\end{tabular}
\end{adjustbox}
}
\end{table}

\section{\filegramengine{}: Behavioral Data Generation}
\label{sec:filegram}

In this section, we introduce \filegramengine{}, the data-generation component of \filegram{} for synthesizing realistic file-system behavioral traces conditioned on specific user profiles and tasks. We illustrate our task formulation in \cref{sec:input_format}, followed by the data engine in \cref{sec:data_engine}, which simulates controlled, long-term trajectories, translating raw tool usage into typed atomic actions paired with rich file-level artifacts, such as \emph{content deltas}---the precise record of what changed in each file, comprising full snapshots for newly created files and patch \texttt{diffs} for edits---and final agent outputs.

\subsection{Profile \& Task Formulation}
\label{sec:input_format}

\smallskip\noindent\textbf{Profile Design.}
\label{sec:profile}
To systematically model user variance, our schema defines 19 fine-grained attributes per profile as detailed in \Cref{sec:appendix_attributes} of Appendix, combining basic semantic identity fields (\emph{e.g.}, role and language) with six core behavioral dimensions shown in \cref{tab:dimensions}. These dimensions capture recurring user patterns across distinct workflows:  \emph{Consumption Pattern} (A), \emph{Production Style} (B), \emph{Organization Preference} (C), \emph{Iteration Strategy} (D), \emph{Curation} (E), and \emph{Cross-Modal Behavior} (F). We discretize each dimension into three distinct tiers -- \texttt{L/M/R}, to capture a realistic spectrum of behavioral styles, ranging from minimalist and rapid execution to exhaustive and structured iteration, thus providing a controlled basis for the benchmark attributes in \filegrambench{}. To ensure this control mechanism yields realistic behaviors, we let human verifiers validate that the generated traces clearly reflect the profile specifications. Building on this validated schema, we instantiate 20 diverse profiles with varying \texttt{L/M/R} combinations. We calibrate evaluation difficulty by pairing profiles at two granularities: we test subtle behavioral shifts by differing in 1--2 dimensions, and macro-level distinctions through pairs differing over 5 dimensions.

\begin{figure*}[!t]
\centering
\includegraphics[width=\linewidth]{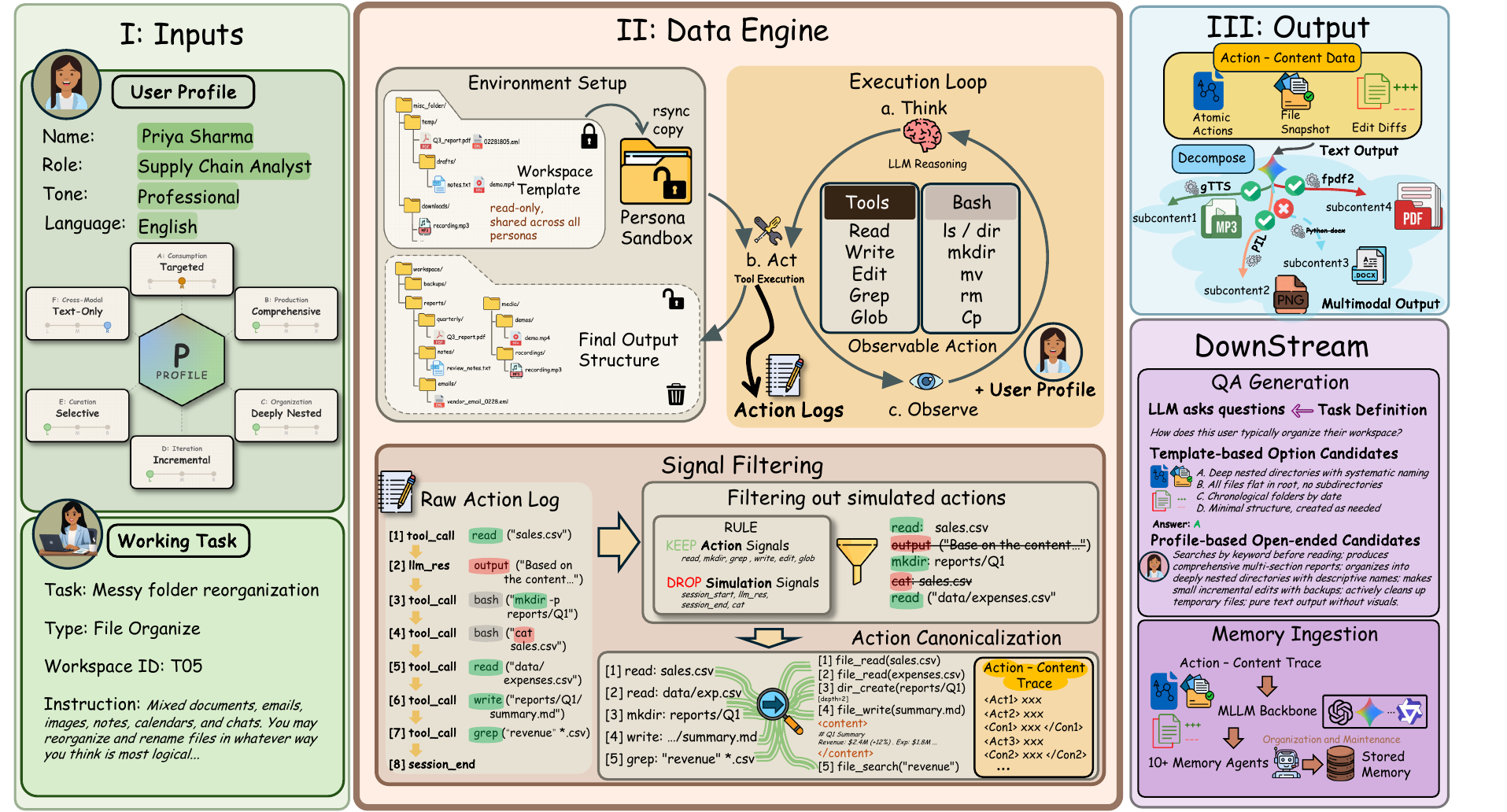}
\caption{\textbf{Data generation pipeline.} \filegramengine{} generates one trajectory per profile--task pair. Agents execute in profile-isolated workspaces for each task; raw tool traces are filtered and canonicalized to retain real action signals while removing simulation artifacts, and outputs are materialized as standardized behavioral traces with aligned text/document/visual views for cross-modal evaluation.}
\label{fig:data_pipeline}
\end{figure*}

\begin{table}[!t]
\caption{Six behavioral dimensions with L/M/R tiers used for profile construction in \filegramengine{}.}
\label{tab:dimensions}
\centering
\resizebox{\linewidth}{!}{
\begin{tabular}{@{}clccc@{}}
\toprule
\textbf{Dim} & \textbf{Name} & \textbf{Left (L)} & \textbf{Middle (M)} & \textbf{Right (R)} \\
\midrule
A & Consumption Pattern & Sequential deep reading & Targeted search-first & Breadth-first browsing \\
B & Production Style & Comprehensive \& detailed & Balanced & Minimal \& concise \\
C & Organization Preference & Deeply nested (3+ levels) & Adaptive (1--2 levels) & Flat (root only) \\
D & Iteration Strategy & Incremental small edits & Balanced refinement & Bulk rewrite \\
E & Curation & Selective (active cleanup) & Pragmatic (moderate cleanup) & Preservative (accumulative) \\
F & Cross-Modal Behavior & Visual-heavy (charts, figures) & Balanced (tables) & Text-only \\
\bottomrule
\end{tabular}
}
\end{table}

\smallskip\noindent\textbf{Task Design.}
\label{sec:task}
We derive task design directly from the six profile dimensions: each task is constructed to elicit trace-observable behavioral signals \cite{krathwohl2002revision}. Tasks are organized into six types---\emph{Understand}, \emph{Create}, \emph{Organize}, \emph{Synthesize}, \emph{Iterate}, and \emph{Maintain}---ranging from focused single-dimension probes to compositional multi-dimension settings. In total, we curate 32 tasks (16 text-centric, 16 multimodal), each initialized with a pre-populated workspace curated from real personal file collections in HippoCamp~\cite{yang2026hippocamp}, collectively comprising 615 diverse input files spanning videos, audio, images, spreadsheets, presentations, and PDFs. Details are in \Cref{sec:appendix_tasks}.

\smallskip\noindent\textbf{Behavioral Perturbation.}
To prevent the generation of unrealistically static personas, we introduce deliberate \emph{behavioral perturbation}. Specifically, for each profile, five trajectories are forced to undergo a localized shift in a task-relevant dimension by a single tier defined in~\cref{tab:dimensions}. This injection of controlled noise serves a dual purpose: first, it mirrors the natural behavioral fluctuations of real-world users, preventing memory systems from exploiting overly consistent, shortcut-style heuristics. Second, these perturbed trajectories establish a critical foundation for \filegrambench{}, explicitly powering Track~3 to evaluate a system's robustness and its capacity for persona drift detection.

\subsection{Data Engine and Composition}
\label{sec:data_engine}

\smallskip\noindent\textbf{\filegramengine{}.} To synthesize the behavioral trajectories, the \filegramengine{} pairs each profile with every task, yielding 640 unique execution combinations as illustrated in \cref{fig:data_pipeline}. Because the tasks inherently alter the file system by creating, editing, and reorganizing files, we make sure each execution occurs within an isolated, sandboxed workspace to strictly prevent behavioral cross-contamination. Within each sandbox, a tool-using agent~\cite{anthropic2025claude_haiku_4_5} prompted with the assigned persona and executes the task through a continuous think--act--observe loop~\cite{yao2022react}. Crucially, the engine organizes these interactions at two distinct levels: raw shell commands and tool calls are abstracted into typed, atomic actions, while each event is systematically paired with its corresponding content delta, capturing full snapshots for new files and precise patch \texttt{diffs} for edits. Finally, a post-execution filter examines artificial simulation traces, such as LLM thought processes and intermediate error logs, ensuring the resulting trajectories contain only pure, behaviorally meaningful file operations.

\begin{figure*}[t]
\centering
\includegraphics[width=\linewidth]{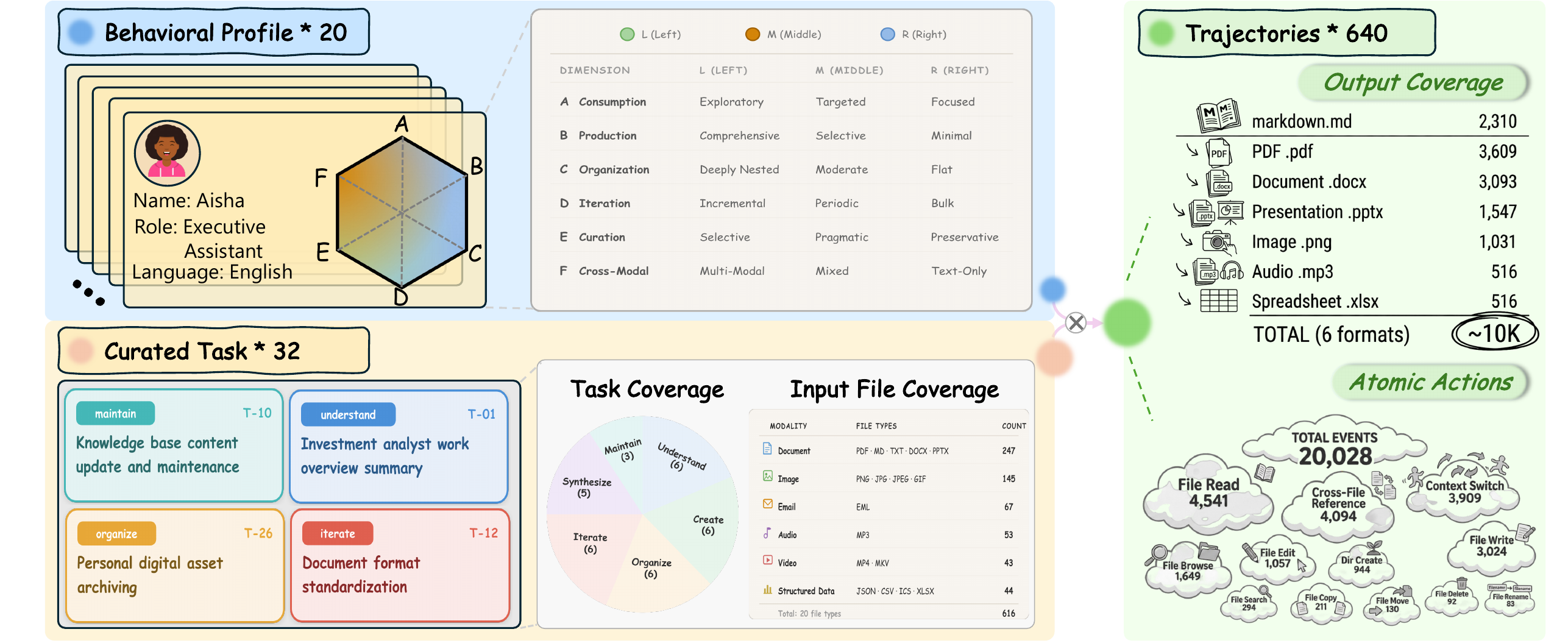}
\caption{\textbf{Data distribution.} 20 profiles $\times$ 32 tasks yield 640 trajectories comprising ${\sim}$10K output files and 20,028 atomic actions.}
\label{fig:data_distribution}
\end{figure*}

\smallskip\noindent\textbf{Dataset Composition.}
\label{sec:output}
The pipeline produces 640 behavioral trajectories, comprising 20,028 atomic actions and approximately 2.5K agent-generated files.
By interleaving structured procedural logs with fine-grained content deltas, each trajectory provides a highly granular chronological record that collaboratively serves as the empirical foundation for \filegrambench{}. To further enrich modality diversity for cross-modal evaluation, we develop a decomposition pipeline that segments text-based outputs into semantically coherent sections and renders them across diverse target modalities. This expansion yields over 10K multimodal files spanning PDFs, slide presentations, images, audio narrations, and other formats (\cref{fig:data_distribution}). Together, these trajectories and their multimodal derivatives constitute the foundational corpus for \filegrambench{} evaluations.

\section{\filegrambench{}: Evaluation Framework}
\label{sec:bench}

\begin{wrapfigure}[19]{R}{0.42\columnwidth}
\centering
\includegraphics[width=\linewidth]{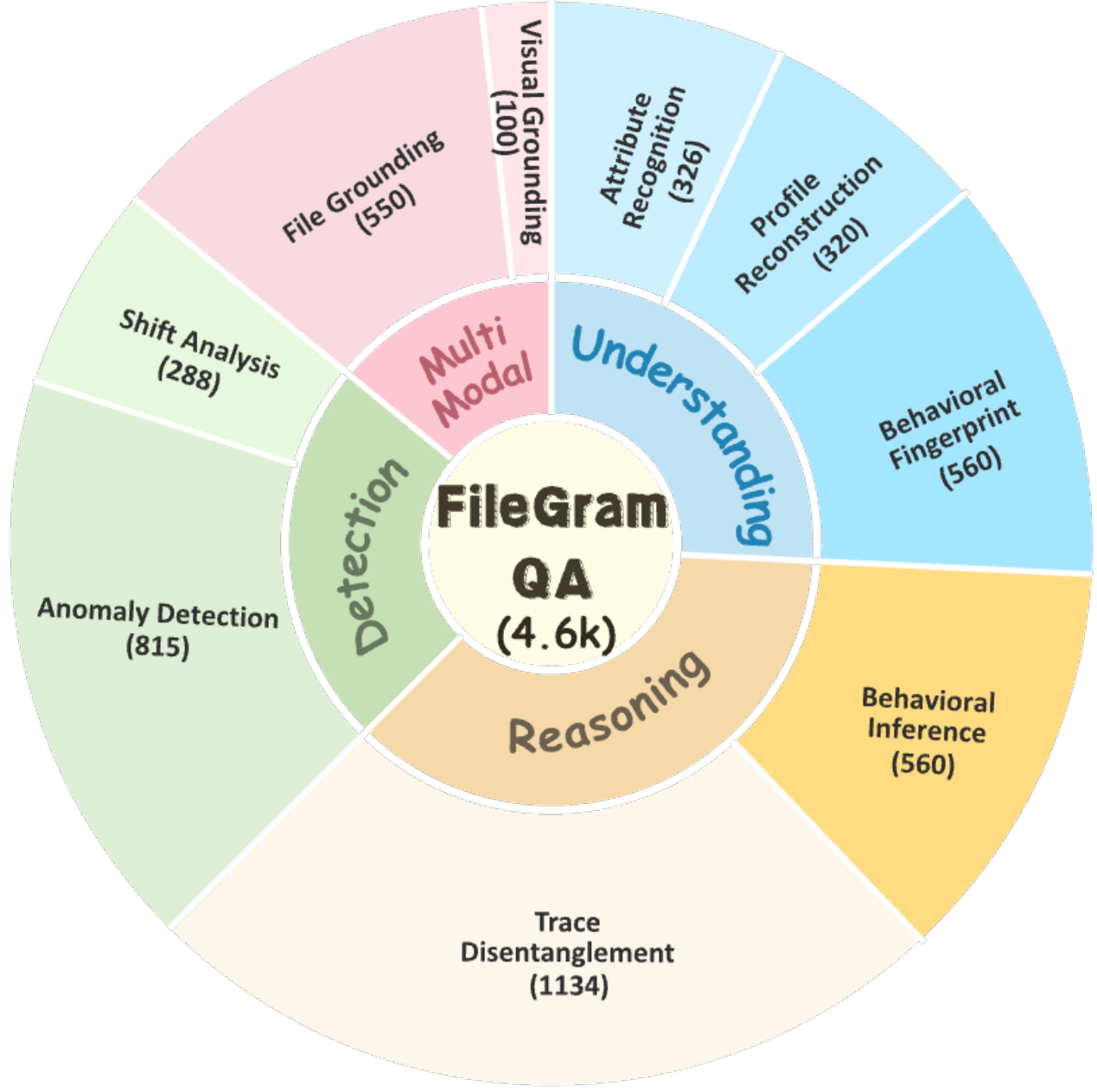}
\caption{\textbf{\filegramqa{} distribution.} 4.6K questions by track (inner) and sub-task (outer).}
\label{fig:filegramqa_corner}
\end{wrapfigure}

\filegrambench{} comprises 4.6K memory-targeted QA pairs across nine sub-tasks organized into four tracks (\cref{fig:filegramqa_corner}), covering procedural, semantic, and episodic memory channels. The benchmark spans both simulated behavioral trajectories and real-world human screen recordings, with ground truth derived from predefined user profiles to ensure objective evaluation.

\subsection{Automatic QA Generation}
\label{sec:qa_generation}

\filegrambench{} converts behavioral trajectories from \filegramengine{} into structured evaluation items through a template-based pipeline.

\smallskip\noindent\textbf{MCQ Construction.}
Answer options are constructed from predefined profile attributes in the templates, with distractors drawn from fine-grained profile pairs differing in only 1--2 dimensions to ensure genuine behavioral discrimination. Trajectory sequence fragments serve as both evidence context and, in some sub-tasks, answer candidates. GPT-4.1 generates the natural-language questions given the options and context.

\smallskip\noindent\textbf{Open-ended Construction.}
Ground-truth answers are derived from profile templates. We define per-attribute rubrics and use an LLM judge to score each response on a Likert 1--5 scale.

\smallskip\noindent\textbf{Real-world Annotation.}
Beyond simulated trajectories, we also collect real-world screen recordings. We first convert simulated trajectories into GUI-level operation sequences as behavioral guidance videos. Human participants then receive the task description, behavioral profile, and this guidance video, and perform the task while their screen is recorded. This pipeline ensures controllable data collection while grounding evaluation in authentic user behavior.

\subsection{QA Taxonomy}
\label{sec:qa_taxonomy}

\smallskip\noindent\textbf{Track 1: Understanding.}
Given $N$ trajectories from one user, recover that user's behavioral profile.
\emph{Attribute Recognition} (326 MCQs, 3-choice): identify the L/M/R tier on a specified behavioral dimension or infer semantic attributes (role, tone, language);
\emph{Behavioral Fingerprint} (560 MCQs, 4-choice): given a single anonymous trajectory, match it to one of four candidate profiles;
\emph{Profile Reconstruction} (free-form): produce a structured assessment across all six behavioral dimensions through 19 user attributes.

\smallskip\noindent\textbf{Track 2: Reasoning.}
Pattern-level inference and disentanglement under ambiguity.
\emph{Behavioral Inference} (560 MCQs, 4-choice): given 31 trajectories with one task held out, predict behavior on the unseen task;
\emph{Trace Disentanglement} (1,134 MCQs, 2--4 choices): given interleaved event streams from two users on the same task, identify the primary behavioral difference.

\smallskip\noindent\textbf{Track 3: Detection.}
Per-session memory under behavioral drift. Using the perturbation design from \cref{sec:task}, 5 of 32 trajectories per profile shift one dimension by one tier.
\emph{Anomaly Detection} (815 MCQs, 5--6 choices): given trajectories mixed with one impostor from a similar profile, identify the impostor session;
\emph{Shift Analysis} (288 MCQs, 3--6 choices): given baseline and one perturbed trajectory, identify which dimension shifted and in which direction.

\begin{figure}[t]
\centering
\includegraphics[width=\columnwidth]{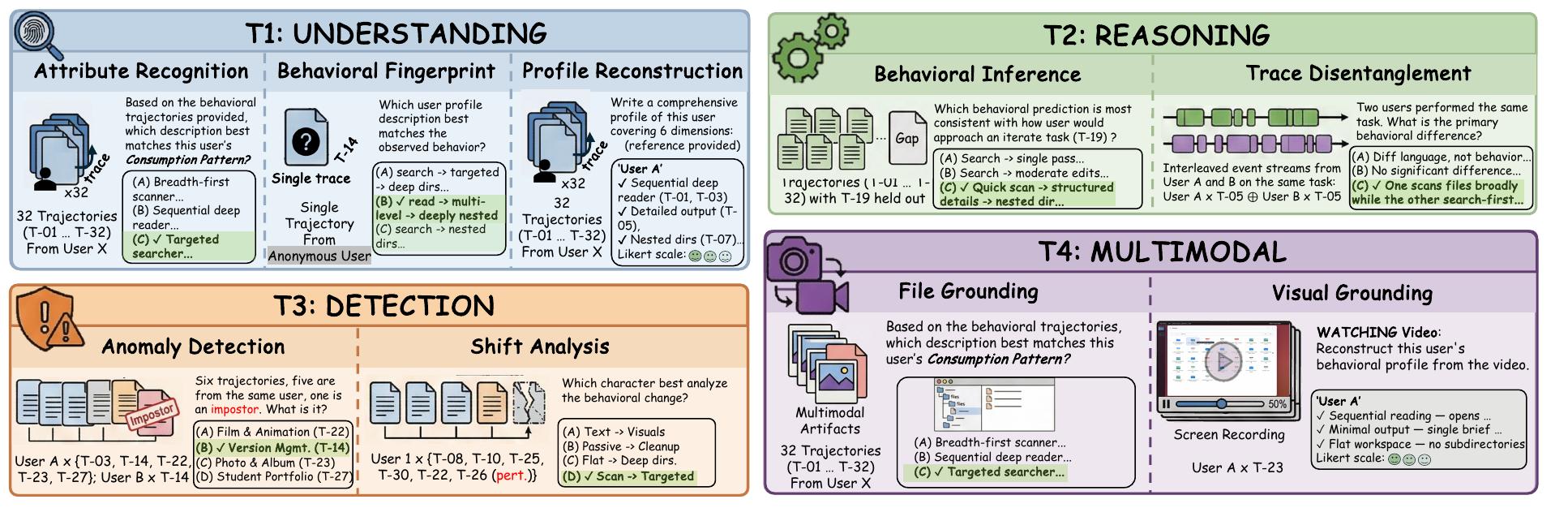}
\caption{\textbf{QA examples from \filegrambench{}.} Representative questions from the four tracks, including both MCQ and open-ended formats.}
\label{fig:qa_example}
\end{figure}

\smallskip\noindent\textbf{Track 4: Multimodal Grounding.}
Extend evaluation to vision-centric setting with rendered documents and real-world screen recordings.
\emph{File Grounding} (550 MCQs): answer the same behavioral questions as Tracks~1--3, but with file outputs presented as rendered PDFs and images instead of raw text;
\emph{Visual Grounding} (100, free-form, real-world): given the first half of a real participant's screen recording, predict subsequent file operations and behavioral patterns.

\smallskip\noindent\textbf{Channel-wise Grouping.}
We map each sub-task to one of three channels: \emph{procedural} for operation-level patterns, \emph{semantic} for content-level understanding, and \emph{episodic} for temporal consistency and drift detection across session.

\subsection{Evaluation Protocol}
\label{sec:settings}

\smallskip\noindent\textbf{Two-stage pipeline and leakage control.}
(1)~\textbf{Ingest}: each method processes raw trajectories (atomic actions + content deltas) using its own memory pipeline. Methods requiring an LLM during ingestion use the same backbone (Gemini 2.5-Flash~\cite{comanici2025gemini}), isolating memory design as the independent variable. (2)~\textbf{Answer}: Gemini 2.5-Flash answers MCQs using only the retrieved memory. To prevent leakage, models never access ground-truth profiles, dimension definitions, or perturbation tags; ingestion is restricted to raw trajectories and content deltas.
\section{\filegramos{}: Bottom-Up Memory Framework}
\label{sec:filegramos}

\begin{figure*}[t]
\centering
\includegraphics[width=\linewidth]{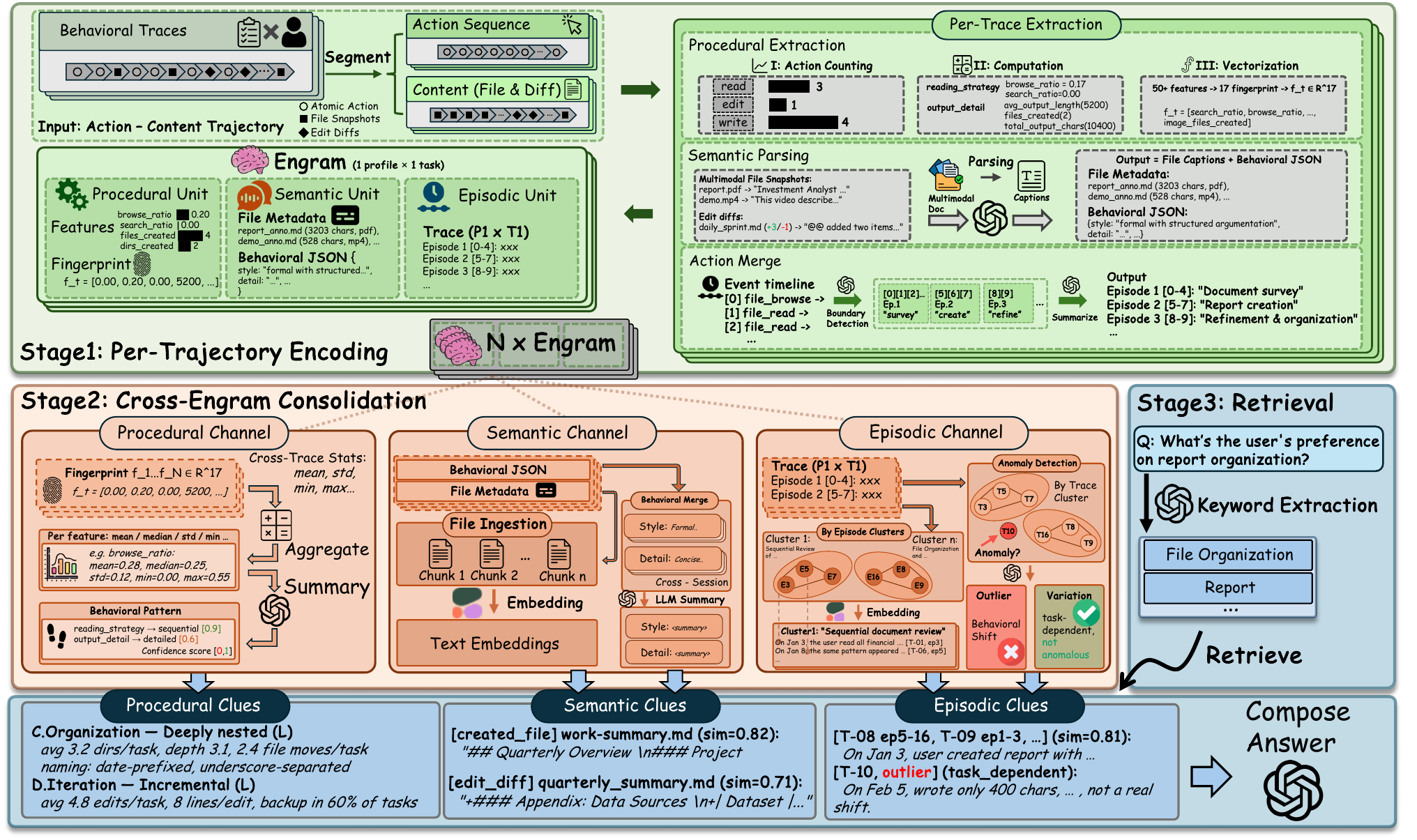}
\caption{\textbf{\filegramos{} architecture.} Three-stage pipeline: (1) per-trajectory encoding of traces via parallel extraction streams into an Engram; (2) cross-engram consolidation routing data into procedural, semantic, and episodic channels, including an LLM verifier for \emph{variation vs.\ outlier}; and (3) query-adaptive retrieval.}
\label{fig:architecture}
\end{figure*}

\filegramos{} is a bottom-up, action-aware memory framework designed for file-centric user behavior. Rather than prematurely summarizing trajectories into free-form narratives, it builds structured memory from raw event traces through three stages: per-trajectory encoding that processes distinct behavioral and semantic streams into an atomic \emph{Engram}, cross-engram consolidation that routes these units into three specialized memory channels, and a lightweight retrieval stage that composes answers from the consolidated clues. Figure~\ref{fig:architecture} summarizes the overall pipeline.

\subsection{Stage 1: Per-Trajectory Encoding}
\label{sec:delta}

The first stage transforms raw, noisy inputs (action sequences and file content/diffs) into a structured atomic memory unit called an \textbf{Engram}. While a raw trajectory is simply a chronological event sequence, an Engram distills it into a compact, multi-faceted representation that jointly captures procedural statistics, semantic content, and episodic structure. To capture the multifaceted nature of user behavior, the data flows through three parallel extraction pipelines:

\smallskip\noindent\textbf{Procedural Extraction.}
This pipeline isolates the mechanics of user actions. It begins with \emph{Action Counting} (e.g., tracking reads, edits, writes), followed by \emph{Computation} to derive higher-level metrics such as browse ratios and average output lengths. Finally, \emph{Vectorization} compresses over 50 behavioral features into a dense 17-dimensional fingerprint $\mathbf{f}_j \in \mathbb{R}^{17}$.

\smallskip\noindent\textbf{Semantic Parsing.}
This stream extracts meaning from the content itself. Multimodal file snapshots (documents, videos) and edit diffs are passed through a vision-language model to generate structural captions and a behavioral descriptor that summarizes the user's style, formatting preferences, and detail level.

\smallskip\noindent\textbf{Action Merge.}
Simultaneously, the raw event timeline undergoes boundary detection to segment continuous traces into discrete, logical episodes (e.g., ``Document survey'' $\rightarrow$ ``Report creation'').

These three streams converge to instantiate an \emph{Engram} for a specific profile and task. Each Engram explicitly stores a \textbf{Procedural Unit} (the vectorized fingerprint), a \textbf{Semantic Unit} (file metadata and behavioral descriptor), and an \textbf{Episodic Unit} (segmented trace episodes).

\subsection{Stage 2: Cross-Engram Consolidation}
\label{sec:channels}

Once $N$ Engrams are generated across multiple sessions, the \textbf{Engram Consolidator} unpacks and routes their components into a unified \textbf{MemoryStore} divided into three complementary channels:

\smallskip\noindent\textbf{Procedural Channel.}
This channel establishes stable behavioral traits by aggregating the 17-D fingerprints ($\mathbf{f}_1 \dots \mathbf{f}_N$) from the Engrams' Procedural Units. It computes cross-trace statistics (mean, median, standard deviation, min, max) for each feature. These aggregated statistics form stable \emph{Procedural Clues}, allowing the system to confidently categorize behaviors like ``Deeply nested organization'' or ``Incremental iteration.''

\smallskip\noindent\textbf{Semantic Channel.}
Taking the behavioral descriptors and file metadata from the Semantic Units, this channel handles content ingestion and embedding. Text data is divided into chunks and embedded to group similar content. An LLM then performs a cross-session summary, merging distinct styles and detail preferences into unified \emph{Semantic Clues}.

\smallskip\noindent\textbf{Episodic Channel.}
This channel maintains temporal fidelity and detects behavioral drift using the Episodic Units. Trajectories are clustered into behavioral modes based on sequence similarity. To detect anomalies, session fingerprints are z-score normalized and evaluated by their distance to the cluster centroid:
\begin{equation}
\label{eq:anomaly}
z_k^{(j)}=\frac{f_k^{(j)}-\mu_k}{\sigma_k+\epsilon},\quad
\delta_j=\|\mathbf z_j-\bar{\mathbf z}\|_2,\quad
\hat y_j=\mathbb{I}\!\left[\delta_j>\mu_\delta+\tau\sigma_\delta\right],
\end{equation}
with $\tau{=}1.5$. Since numeric outliers in file-centric tasks are often intentional, flagged sessions are passed to an LLM-based \textbf{Anomaly Judge}:
\begin{equation}
\label{eq:verify}
r_j=\mathrm{LLM}(\psi_j)\in\{\texttt{variation},\texttt{outlier},\texttt{uncertain}\}.
\end{equation}
This explicitly disambiguates task-dependent \emph{variations} from genuine behavioral \emph{shifts}, outputting contextual \emph{Episodic Clues}.

\subsection{Stage 3: Query-Adaptive Retrieval}
\label{sec:deferred}

By maintaining three distinct channels, \filegramos{} defers the final interpretation of the memory until query time. Given a user query, the system performs \textbf{Keyword Extraction} to identify the target dimension, e.g., ``File Organization''. It then adaptively retrieves the pre-computed clues from the MemoryStore---pulling structural habits from the Procedural Clues, stylistic preferences from Semantic Clues, and flagged deviations from Episodic Clues---and routes them to a final LLM generation step to compose a grounded, evidence-backed answer.

\section{Experiments}
\label{sec:exp}

\subsection{Setup}
\label{sec:setup}

\smallskip\noindent\textbf{Data and Input.}
We evaluate 640 trajectories from \filegramengine{} under three settings. The \emph{Text} setting uses original Markdown agent outputs. The \emph{Multimodal} setting renders outputs as PDFs and images. The \emph{Real-World} setting replaces simulated traces with human screen recordings. Behavioral event logs remain identical across the first two settings, and we utilize Gemini 2.5-Flash as the shared video captioner across the three settings.

\smallskip\noindent\textbf{Methods.}
\label{sec:baselines}
We evaluate \filegramos{} against 12 methods using Gemini 2.5-Flash \cite{comanici2025gemini} as the shared QA backbone. The baselines fall into three distinct groups: (1) context methods including Full Context, Naive RAG, and VisRAG~\cite{yu2025visragvisionbasedretrievalaugmentedgeneration}, (2) text interaction memory methods spanning Mem0~\cite{chhikara2025mem0}, Zep~\cite{rasmussen2025zep}, MemOS~\cite{li2025memos}, EverMemOS~\cite{hu2026evermemos}, and SimpleMem~\cite{liu2026simplemem}, and (3) multimodal memory methods featuring MMA~\cite{lu2026mma} and MemU~\cite{lee2025memu}.

\subsection{Results Analysis}
\label{sec:main_results}

\begin{table*}[t]
\caption{\textbf{Main results on \filegrambench{}.}
All scores are accuracy (\%) scaled 0--100. $^\dagger$Open-ended sub-tasks are scored by LLM judge (Likert 1--5, rescaled). $^\ddagger$Multimodal memory with native non-text ingestion. Sub-task and token definitions are detailed below the table.}
\label{tab:main}
\centering
\renewcommand{\arraystretch}{1.15}
\setlength{\tabcolsep}{2pt}
\begin{adjustbox}{max width=\textwidth}
\begin{threeparttable}
\begin{tabular}{@{}l|cc|ccc|cc|cc|cc|ccc|c@{}}
\toprule
\multirow{2}{*}{\textbf{Method}}
& \multicolumn{2}{c|}{\textbf{Tokens}}
& \multicolumn{3}{c}{\textbf{T1: Understanding}}
& \multicolumn{2}{c}{\textbf{T2: Reasoning}}
& \multicolumn{2}{c}{\textbf{T3: Detection}}
& \multicolumn{2}{c|}{\textbf{T4: MM}}
& \multicolumn{3}{c|}{\textbf{Channel}}
& \multirow{2}{*}{\textbf{Avg}}
\\
\cmidrule{2-3} \cmidrule(lr){4-6} \cmidrule(lr){7-8} \cmidrule(lr){9-10} \cmidrule(lr){11-12} \cmidrule{13-15}
& \textbf{In.} & \textbf{Out.}
& \makecell[c]{\textbf{Attr}\\\textbf{Rec}} & \makecell[c]{\textbf{Behav}\\\textbf{FP}} & \makecell[c]{\textbf{Prof}\\\textbf{Rec}\rlap{$^\dagger$}}
& \makecell[c]{\textbf{Behav}\\\textbf{Inf}} & \makecell[c]{\textbf{Trace}\\\textbf{Dis}}
& \makecell[c]{\textbf{Anom}\\\textbf{Det}} & \makecell[c]{\textbf{Shift}\\\textbf{Ana}}
& \makecell[c]{\textbf{File}\\\textbf{Grd}} & \makecell[c]{\textbf{Vis}\\\textbf{Grd}{$^\dagger$}}
& \textbf{Proc} & \textbf{Sem} & \textbf{Epi}
& \\
\midrule
No Context      & \textemdash & \textemdash & 36.2 & 25.7 & -- & 17.4 & 36.9 & 19.0 & 20.5 & 23.8 & -- & 25.7 & 38.2 & 19.7 & 25.4 \\
\midrule
Full Context    & 625.2K & 45.9K & 40.5 & 31.1 & 50.0 & 30.6 & \underline{80.5} & 36.8 & 37.8 & 42.5 & 7.0 & 50.7 & 43.0 & 37.3 & 48.0 \\
Naive RAG       & 625.2K & 3.9K & 48.2 & 27.7 & 46.8 & 26.4 & 64.1 & 38.4 & 20.1 & 35.1 & 5.5 & 42.0 & 49.3 & 29.7 & 40.5 \\
Eager Summ.     & 625.2K & 3.7K & 45.1 & 29.6 & 55.6 & \underline{39.3} & 65.9 & 59.7 & 36.1 & 44.0 & 6.5 & 49.8 & 49.3 & 48.4 & 49.5 \\
VisRAG~\cite{yu2025visragvisionbasedretrievalaugmentedgeneration}$^\ddagger$          & 609.8K & 10.0K & \textbf{53.4} & \underline{33.2} & 56.3 & 32.9 & 72.8 & 64.5 & 35.4 & \underline{45.3} & 7.0 & \underline{51.2} & \textbf{55.2} & 54.7 & \underline{51.9} \\
\midrule
Mem0~\cite{chhikara2025mem0}            & 119.9K & 3.0K & 44.2 & 26.4 & 48.1 & 21.4 & 50.4 & 23.8 & 28.5 & 29.5 & 4.0 & 33.4 & 47.8 & 26.0 & 33.2 \\
Zep~\cite{rasmussen2025zep}             & 219.1K & 3.8K & 43.6 & 28.4 & 50.4 & 27.4 & 61.0 & 37.5 & 28.1 & 35.4 & 5.0 & 41.3 & 44.6 & 33.0 & 40.2 \\
MemOS~\cite{li2025memos}           & 302.3K & 4.2K & 44.2 & 24.8 & 52.0 & 23.0 & 57.3 & 26.3 & 28.1 & 32.0 & 4.5 & 37.2 & 47.2 & 27.2 & 36.2 \\
SimpleMem~\cite{liu2026simplemem}       & 9.3K & 3.5K & 43.6 & 20.2 & \underline{56.6} & 28.2 & 47.5 & 21.8 & 28.5 & 29.0 & 4.5 & 33.3 & 47.2 & 26.2 & 32.9 \\
EverMemOS~\cite{hu2026evermemos}       & 1098.9K & 8.4K & 48.8 & 30.2 & \textbf{57.7} & \underline{39.3} & 62.2 & \textbf{71.4} & \textbf{38.9} & 44.5 & \underline{7.5} & 48.7 & 50.8 & \underline{55.9} & 49.9 \\
MemU~\cite{lee2025memu}$^\ddagger$            & 293.6K & 7.9K & 47.9 & 27.3 & 50.4 & 30.4 & 65.7 & 46.0 & 33.0 & 39.8 & 6.0 & 44.9 & 49.3 & 39.8 & 44.4 \\
MMA~\cite{lu2026mma}$^\ddagger$             & 331.8K & 4.5K & \underline{51.2} & 29.8 & 51.8 & 28.9 & 57.4 & 57.5 & 32.6 & 41.3 & 5.5 & 42.8 & 51.6 & 53.1 & 44.7 \\
\midrule
\rowcolor{oursrow}
\filegramos{}   & 109.7K & 4.3K & 50.6 & \textbf{35.2} & 54.2 & \textbf{42.1} & \textbf{80.9} & \underline{70.2} & \underline{37.8} & \textbf{55.8} & \textbf{8.5} & \textbf{60.1} & \underline{54.6} & \textbf{58.9} & \textbf{59.6} \\
\bottomrule
\end{tabular}

\begin{tablenotes}[flushleft]
            \scriptsize
            \item \textit{T1} --- \textbf{AttrRec}: Attribute Recognition (326, 3-choice); \textbf{BehavFP}: Behavioral Fingerprint (\fpCount{}, 4-choice); \textbf{ProfRec}: Profile Reconstruction (\profRecCount{}, free-form).
            \item \textit{T2} --- \textbf{BehavInf}: Behavioral Inference (\cfCount{}, 4-choice); \textbf{TraceDis}: Trace Disentanglement (1,134, 2--4 choices).
            \item \textit{T3} --- \textbf{AnomDet}: Anomaly Detection (815, 5--6 choices); \textbf{ShiftAna}: Shift Analysis (288, 3--6 choices).
            \item \textit{T4} --- \textbf{FileGrd}: File Grounding (\fileGrdCount{}, mixed); \textbf{VisGrd}: Visual Grounding (\visGrdCount{}, free-form). Text methods use a shared text parser; VisRAG and $^\ddagger$methods use native multimodal input.
            \item \textit{Channel} --- \textbf{Proc}: procedural; \textbf{Sem}: semantic; \textbf{Epi}: episodic. \textit{Tokens} --- \textbf{In.}: total stored memory per profile (avg.\ over 20 profiles); \textbf{Out.}: retrieved context per query (avg.).
        \end{tablenotes}
\end{threeparttable}
\end{adjustbox}
\end{table*}

\begin{figure*}[!t]
\centering
\includegraphics[width=\linewidth]{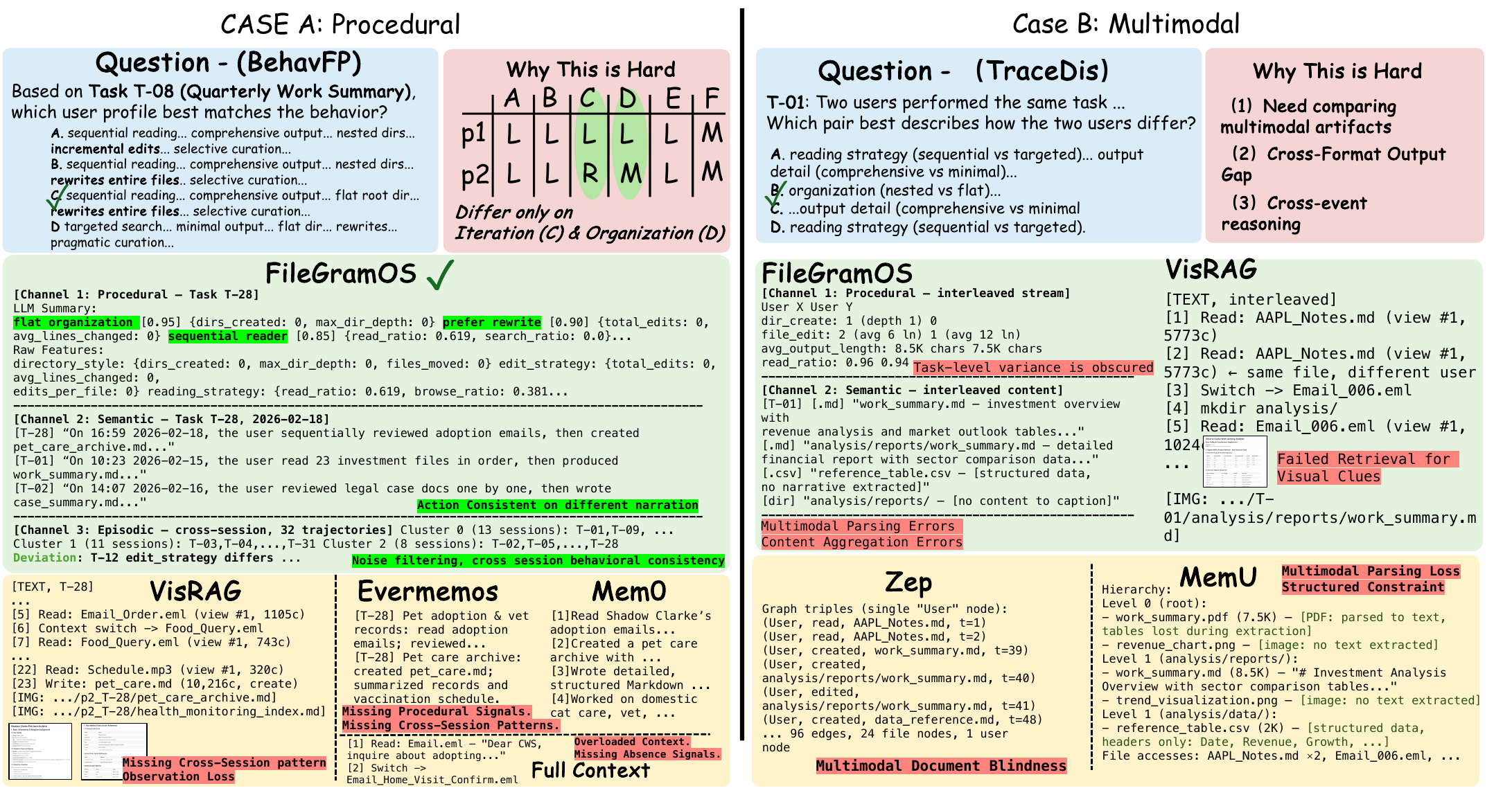}
\caption{\textbf{Qualitative comparison.} \emph{Left:} A BehavFP question where \filegramos{}'s three-channel architecture---procedural statistics, semantic narration, and episodic clustering---jointly recover the correct profile, while baselines each miss different signals. \emph{Right:} A TraceDis question involving multimodal artifacts, where cross-format output gaps and parsing losses cause widespread failures.}
\label{fig:qualitative}
\end{figure*}

\smallskip\noindent\textbf{Bottom-up Structure Surpasses Narrative Summarization.}
Among memory-targeted methods, \filegramos{} scores 59.6\%, significantly outperforming the strongest narrative baseline EverMemOS at 49.9\%. The core advantage lies in abstraction timing. Narrative-first methods like Mem0, Zep, MemOS, SimpleMem, and EverMemOS summarize trajectories during ingestion. This prematurely erases key behavioral discriminators like action counts, directory depth, and edit granularity. The result is systematic flattening where distinct profiles receive identical generic descriptors, like ``structured'', ``methodical'', and ``comprehensive'', despite differing operations. In contrast, as shown in \cref{fig:qualitative}, \filegramos{} prevents this by preserving distributional statistics at ingest time and deferring semantic abstraction until query time.

\smallskip\noindent\textbf{Track and Channel Overview.}
Track~1 and 2 are partially solvable, while Track~3 exposes a clear detection-vs-explanation gap. Specifically, (1) AnomDet evaluates cross-session summarization: methods that aggregate behavioral norms across trajectories, such as EverMemOS and \filegramos{}, achieve over 70\% accuracy, whereas flat memory systems like Mem0 and SimpleMem remain near random at 21--26\%; (2) ShiftAna examines trace perturbations along a single behavioral dimension entangled with normal cross-task variance. While existing models can detect overall deviations, they fail to attribute these changes to specific dimensions or directions. In contrast, \filegramos{} surpasses baselines by leveraging channel-wise procedural cues, while the semantic track remains more competitive: VisRAG and EverMemOS rival \filegramos{} by effectively capturing formatting and content information. These results demonstrate that fine-grained operational micro-structure serves as the decisive signal for file-system personalization, whereas semantic understanding can often be approximated through conventional retrieval or summarization strategies.

\smallskip\noindent\textbf{Context Baselines vs.\ Memory-targeted Methods.}
The naive Full Context approach outperforms dedicated memory pipelines by simply concatenating raw events. This strategy proves that preserving complete evidence occasionally outweighs semantic abstraction. The TraceDis task clearly demonstrates this advantage, as Full Context achieves a score similar to \filegramos{} by directly comparing complete action chains. Meanwhile, narrative methods fall far behind because their summaries discard critical sequential diversity signals. However, raw concatenation fails on tasks demanding cross-session comparison. Full Context drops significantly below \filegramos{} in these scenarios, because detecting outliers across 32 trajectories strictly requires structured aggregation. Beyond text methods, vision-augmented retrieval provides a distinct alternative. VisRAG dominates the Semantic channel by leveraging page-image retrieval to capture layout cues. However, it still lacks the behavioral abstraction required for procedural tasks, causing a substantial performance drop in those areas.

\smallskip\noindent\textbf{Multimodal Memory Methods.}
Multimodal memory systems like MMA and MemU fail to outperform the strongest text-only baselines. While mechanisms like confidence-scored retrieval and VLM captioning assist with specific anomaly detection tasks or non-text ingestion, they do not yield stronger overall behavioral discrimination. \filegramos{} surpasses these methods by a wide margin, proving that simply handling multimodal inputs is insufficient. The critical factor remains how behavioral evidence is structured and preserved. Furthermore, rendered page images are inherently blind to operation-level statistics such as file counts, output lengths, and edit frequencies, as well as file-system structures like directory depth and naming conventions. This limitation causes all vision-based methods to fail on these dimensions even when they succeed on formatting cues, as illustrated in \cref{fig:qualitative}.

\smallskip\noindent\textbf{Multimodal and real-world gap.}
When transitioning to the Multimodal setting where outputs become PDFs or images, text-only methods fall toward baseline levels. While MemU mitigates this decline through VLM captioning, \filegramos{} demonstrates the highest resilience because its procedural channel relies on modality-invariant event logs. Moving to the Real-World setting widens this performance gap even further. Specifically, accuracy on human screen recordings drops to single digits across all evaluated methods. This sharp decline reveals a substantial distance between structured trace analysis and actual video-level behavioral understanding. The primary reason for this struggle is that simulated trajectories provide clean action logs, whereas real-world recordings introduce noise, variable pacing, and unstructured visual input. Consequently, this unexplored sim-to-real gap alongside persistent difficulties in shift attribution and open-ended profile reconstruction defines concrete research frontiers for future memory-centric personalized systems.

\section{Conclusion}
\label{sec:conclusion}

In this paper, we present the unified \filegram{} framework, encompassing \filegramengine{} for trajectory generation, \filegrambench{} for diagnostic evaluation, and \filegramos{} as a bottom-up reference method, to make file-system behavioral personalization measurable and reproducible. Using this framework, we conduct extensive evaluations, highlighting significant challenges within this domain. Shared workspace content provides weak personalization signals compared to operation-level traces, meaning early narrative summarization inadvertently flattens distinct user behaviors. Furthermore, shift attribution remains a critical bottleneck because systems can easily detect anomalies but consistently fail to explain the exact nature and direction of those behavioral changes. Together, we hope the framework, along with the exposed challenges, will pave the way for developing personalized memory-centric file-system agents.

\clearpage

\bibliographystyle{assets/plainnat}
\bibliography{main}

\newpage
\beginappendix

This supplementary material is organized into five parts.
\Cref{sec:appendix_related} extends the related work.
\Cref{sec:appendix_system} details the \filegramos{} architecture and pipeline.
\Cref{sec:appendix_bench} describes benchmark construction and data.
\Cref{sec:appendix_experiments} presents extended experiments and analysis.
\Cref{sec:appendix_discussion} addresses deployment, ethics, and reproducibility.

\section{Extended Related Work}
\label{sec:appendix_related}

\subsection{Memory Framework Comparison}
\label{sec:appendix_memory}

Beyond the benchmark-level comparison in the main paper, we position \filegramos{} within the broader landscape of memory architectures in \cref{tab:memory-systems}.

\smallskip\noindent\textbf{Dialogue-based and flat-store systems.}
First-generation memory frameworks---MemGPT~\cite{packer2023memgpt}, Mem0~\cite{chhikara2025mem0}, SimpleMem~\cite{liu2026simplemem}---extract semantic facts from dialogue and store them in flat or hierarchical key--value stores.
While effective for conversational recall, they lack procedural modeling and cannot ingest non-textual behavioral evidence.
Graph-based extensions such as HippoRAG~\cite{hipporag}, Zep~\cite{rasmussen2025zep}, and A-MEM~\cite{xu2025amem} introduce relational structure via knowledge graphs and personalized retrieval, yet remain anchored to dialogue as the sole input modality.

\smallskip\noindent\textbf{Multimodal and video memory.}
Recent systems broaden the input space beyond text.
MMA~\cite{lu2026mma} and MemU~\cite{lee2025memu} incorporate vision-language perception; VideoRAG~\cite{videorag}, HippoMM~\cite{lin2025hippomm}, M3-Agent~\cite{long2025m3agent}, and EventMemAgent~\cite{wen2026eventmemagent} process video or audio streams.
Among these, only EventMemAgent partially models procedural behavior through event-level annotations, and none ingests file-system traces.

\smallskip\noindent\textbf{Trajectory-aware and hierarchical designs.}
MemOS~\cite{li2025memos} and EverMemOS~\cite{hu2026evermemos} improve temporal organization through hierarchical consolidation, while Memp~\cite{fang2025memp} is the first to model procedural memory from agent trajectories---but it covers neither semantic nor episodic channels.
Across all families, no existing system jointly covers procedural, semantic, and episodic channels from file-system evidence.

\smallskip\noindent\textbf{Structured-profile and ontology-driven memory.}
CAIM~\cite{caim} organizes user knowledge through an ontology-driven tagging scheme, mapping each interaction to a domain taxonomy before storage; this top-down design contrasts with \filegramos{}'s bottom-up approach, where behavioral dimensions emerge from trace statistics rather than a pre-defined ontology.
O-Mem~\cite{omem} introduces a multi-store persona memory with separate working, short-term, and long-term stores---an architecture that parallels \filegramos{}'s three-channel separation but operates on dialogue turns rather than file-system actions.
More broadly, surveys on agent memory taxonomies advocate distinct procedural, episodic, and semantic substrates---a classification that directly motivates \filegramos{}'s channel design.
Knowledge-graph (KG) based approaches~\cite{rasmussen2025zep,xu2025amem} structure memories as entity--relation triples, enabling traversal-based retrieval; \filegramos{}'s procedural channel serves a related role through aggregate statistics (17-D fingerprints), trading graph flexibility for deterministic reproducibility and higher retrieval efficiency.

\begin{table}[t!]
  \caption{\textbf{Memory framework comparison.} \filegramos{} is the first system to ingest file-system behavioral traces and jointly model procedural, semantic, and episodic channels; all prior systems operate on dialogue or video. Src.: primary input source; MM: multimodal support; Str.: storage structure---Flat, Graph, or Hierarchical; Cons.: temporal consolidation.}
  \label{tab:memory-systems}
  \vspace{2pt}
  \centering
  \renewcommand{\arraystretch}{1.08}
  \setlength{\tabcolsep}{3.0pt}
  {\scriptsize
  \begin{adjustbox}{max width=\columnwidth}
  \begin{tabular}{@{}l c c c c c c c@{}}
  \toprule
  \multirow{2}{*}{\textbf{System}}
    & \multirow{2}{*}{\textbf{Src.}}
    & \multirow{2}{*}{\textbf{MM}}
    & \multirow{2}{*}{\textbf{Str.}}
    & \multicolumn{3}{c}{\textbf{Memory Channel}}
    & \multirow{2}{*}{\textbf{Cons.}} \\
  \cmidrule(lr){5-7}
  & & & & \textbf{Proc.} & \textbf{Sem.} & \textbf{Epi.} & \\
  \midrule
  MemGPT~\cite{packer2023memgpt}
    & Dialogue & \xmark &Hier.
    & \xmark & \cmark & \pmark & \cmark \\
  Mem0~\cite{chhikara2025mem0}
    & Dialogue & \xmark &Flat/Graph
    & \xmark & \cmark & \xmark & \cmark \\
  Zep~\cite{rasmussen2025zep}
    & Dialogue & \xmark &Graph
    & \xmark & \cmark & \cmark & \cmark \\
  A-MEM~\cite{xu2025amem}
    & Dialogue & \xmark &Graph
    & \xmark & \cmark & \cmark & \cmark \\
  HippoRAG~\cite{hipporag}
    & Dialogue & \xmark &Graph
    & \xmark & \cmark & \xmark & \cmark \\
  SimpleMem~\cite{liu2026simplemem}
    & Dialogue & \xmark &Hier.
    & \xmark & \cmark & \cmark & \cmark \\
  MemU~\cite{lee2025memu}
    & Multimodal & \cmark &Hier.
    & \xmark & \cmark & \pmark & \xmark \\
  MMA~\cite{lu2026mma}
    & Multimodal & \cmark &Flat
    & \xmark & \cmark & \pmark & \xmark \\
  VideoRAG~\cite{videorag}
    & Video & \cmark &Graph
    & \xmark & \cmark & \cmark & \cmark \\
  VimRAG~\cite{vimrag}
    & Multimodal & \cmark &Graph
    & \xmark & \xmark & \cmark & \pmark \\
  HippoMM~\cite{lin2025hippomm}
    & Audio+Video & \cmark &Hier.
    & \xmark & \cmark & \cmark & \cmark \\
  M3-Agent~\cite{long2025m3agent}
    & Audio+Video & \cmark &Graph
    & \xmark & \cmark & \cmark & \cmark \\
  EventMemAgent~\cite{wen2026eventmemagent}
    & Video & \cmark &Hier.
    & \pmark & \cmark & \cmark & \cmark \\
  Memp~\cite{fang2025memp}
    & Trajectory & \xmark &Hier.
    & \cmark & \xmark & \xmark & \cmark \\
  MemOS~\cite{li2025memos}
    & Dialogue & \xmark &Hier.
    & \pmark & \cmark & \pmark & \cmark \\
  EverMemOS~\cite{hu2026evermemos}
    & Dialogue & \xmark &Hier.
    & \xmark & \cmark & \cmark & \cmark \\
  \midrule
  \rowcolor{oursrow}
  \textbf{\filegramos{} (Ours)}
    & \textbf{File Sys.} & \cmark & \textbf{Hier.}
    & \cmark & \cmark & \cmark & \cmark \\
  \bottomrule
  \end{tabular}
  \end{adjustbox}
  }
  \vspace{2pt}
\end{table}

\section{System Architecture and Implementation}
\label{sec:appendix_system}


\subsection{Behavioral Signal Schema}
\label{sec:appendix_signal_schema}

Each trajectory is composed of typed \emph{atomic actions} paired with their corresponding \emph{content deltas}, stored as a timestamped sequence in \texttt{events.json}.
\Cref{tab:behavioral_events} catalogues all 22 raw event types: 12 atomic actions retained after cleaning and 10 simulation metadata types that are stripped (74.3\% of all raw events).
In addition, three per-event fields---\texttt{message\_id}, \texttt{model\_provider}, and \texttt{model\_name}---are removed from all retained events, as they leak which LLM engine generated the trajectory rather than reflecting user behavior.

\begin{table*}[ht!]
\centering
\caption{\textbf{Event types in raw trajectories.} The upper block lists the 12 atomic action types retained after cleaning with their total counts across 640 trajectories. The lower block lists the 10 simulation metadata types removed, which account for 74.3\% of all raw events.}
\label{tab:behavioral_events}
\label{tab:simulation_events}
\setlength{\tabcolsep}{6pt}
\resizebox{\linewidth}{!}{
\begin{tabular}{@{}lrcl@{}}
\toprule
\multicolumn{4}{@{}l}{\textbf{Retained: Atomic Actions}} \\
\midrule
\textbf{Event Type} & \textbf{Count} & \textbf{Category} & \textbf{Key Fields} \\
\midrule
\texttt{file\_read}    & 4{,}541 & Read   & path, type, depth, view\_count, view\_range, length, revisit\_ms \\
\texttt{file\_browse}  & 1{,}649 & Read   & dir\_path, files\_listed, depth \\
\texttt{file\_search}  &   294 & Read   & search\_type, query, files\_matched, files\_opened \\
\texttt{file\_write}   & 3{,}024 & Write  & path, type, operation, length, before/after\_hash, media\_ref \\
\texttt{file\_edit}    & 1{,}057 & Write  & path, tool, lines\_added/deleted/modified, diff, before/after\_hash \\
\texttt{dir\_create}   &   944 & Org.   & dir\_path, depth, sibling\_count \\
\texttt{file\_copy}    &   211 & Org.   & src\_path, dest\_path, is\_backup \\
\texttt{file\_move}    &   130 & Org.   & old\_path, new\_path, dest\_depth \\
\texttt{file\_delete}  &    92 & Org.   & path, file\_age\_ms, was\_temporary \\
\texttt{file\_rename}  &    83 & Org.   & old\_path, new\_path, naming\_pattern \\
\texttt{cross\_file\_ref} & 4{,}094 & Flow   & src\_file, target\_file, ref\_type, interval\_ms \\
\texttt{context\_switch}       & 3{,}909 & Flow   & from\_file, to\_file, trigger, switch\_count \\
\cmidrule(r){1-2}
\textbf{Subtotal} & \textbf{20{,}028} & & \\
\midrule
\multicolumn{4}{@{}l}{\textbf{Removed: Simulation Metadata}} \\
\midrule
\texttt{tool\_call}            & 15{,}301 & Sim. & Raw tool invocation log \\
\texttt{llm\_response}         & 13{,}096 & Sim. & LLM token counts, latency, stop reason \\
\texttt{iteration\_start}      & 13{,}096 & Sim. & Agent loop iteration begin marker \\
\texttt{iteration\_end}        & 13{,}096 & Sim. & Agent loop iteration end marker \\
\texttt{fs\_snapshot}           & 1{,}280 & Sim. & Directory tree snapshot at session boundaries \\
\texttt{session\_start}        &    640 & Sim. & Session bookkeeping \\
\texttt{session\_end}          &    640 & Sim. & Session totals \\
\texttt{error\_encounter}      &    233 & Sim. & Infrastructure errors \\
\texttt{error\_response}       &    215 & Sim. & Automatic retry of tool failures \\
\texttt{compaction\_triggered}  &    214 & Sim. & Context window compression \\
\cmidrule(r){1-2}
\textbf{Subtotal} & \textbf{57{,}811} & & \\
\bottomrule
\end{tabular}
}
\end{table*}


\subsection{Procedural Fingerprint Specification}
\label{sec:appendix_fingerprint}

The procedural fingerprint $\mathbf{f}_j \in \mathbb{R}^{17}$ compresses each trajectory's behavioral events into a fixed-length vector spanning all six profile dimensions.
\Cref{tab:fingerprint_features} enumerates the 17 features with their computation rules and source event types.

\begin{table*}[ht!]
\centering
\caption{\textbf{Procedural fingerprint specification.} All 17 features are computed deterministically from cleaned atomic actions, grouped by the six behavioral dimensions. Source event types follow \cref{tab:behavioral_events}.}
\label{tab:fingerprint_features}
\setlength{\tabcolsep}{6pt}
\resizebox{\linewidth}{!}{
\begin{tabular}{@{}lllll@{}}
\toprule
\textbf{Group} & \textbf{Key} & \textbf{Source Events} & \textbf{Computation} & \textbf{Interpretation} \\
\midrule
\multirow{3}{*}{reading\_strategy} & search\_ratio & file\_search, file\_read, file\_browse & $\frac{|\text{search}|}{|\text{read}|+|\text{browse}|+|\text{search}|}$ & Targeted search vs.\ sequential browsing \\
 & browse\_ratio & file\_browse, file\_read, file\_search & $\frac{|\text{browse}|}{|\text{read}|+|\text{browse}|+|\text{search}|}$ & Directory-level exploration \\
 & revisit\_ratio & file\_read & $\frac{|\{e : \text{view\_count}>1\}|}{|\text{read}|}$ & Re-reading previously viewed files \\
\midrule
\multirow{3}{*}{output\_detail} & avg\_output\_length & file\_write: create & $\text{mean}(\text{content\_length})$ & Average verbosity of created files \\
 & files\_created & file\_write: create & $|\text{creates}|$ & Number of output files produced \\
 & total\_output\_chars & file\_write: create & $\sum \text{content\_length}$ & Total production volume \\
\midrule
\multirow{3}{*}{directory\_style} & dirs\_created & dir\_create & $|\text{dir\_create}|$ & Active directory structuring \\
 & max\_dir\_depth & dir\_create & $\max(\text{depth})$ & Deepest nesting level \\
 & files\_moved & file\_move & $|\text{file\_move}|$ & Reorganization via relocation \\
\midrule
\multirow{3}{*}{edit\_strategy} & total\_edits & file\_edit & $|\text{file\_edit}|$ & Post-creation modification frequency \\
 & avg\_lines\_changed & file\_edit & $\text{mean}(\text{added}+\text{deleted})$ & Average edit magnitude \\
 & small\_edit\_ratio & file\_edit & $\frac{|\{e : \Delta\text{lines}<10\}|}{|\text{edits}|}$ & Fraction of incremental refinements \\
\midrule
\multirow{2}{*}{version\_strategy} & total\_deletes & file\_delete & $|\text{file\_delete}|$ & Curation aggressiveness \\
 & delete\_to\_create & file\_delete, file\_write & $\frac{|\text{deletes}|}{|\text{creates}|}$ & Curation vs.\ accumulation \\
\midrule
\multirow{3}{*}{cross\_modal} & structured\_files & file\_write: create & $|\{e : \text{ext} \in \mathcal{S}\}|$ & Structured formats: csv, json, xlsx, etc. \\
 & md\_table\_rows & file\_write: create & $\sum |\texttt{/\^{}|.*|/gm}|$ & Inline tabular content in Markdown \\
 & image\_files & file\_write: create & $|\{e : \text{ext} \in \mathcal{I}\}|$ & Visual content: png, jpg, svg, gif \\
\bottomrule
\end{tabular}
}
\end{table*}

\noindent\textbf{Normalization and consolidation.}
During cross-engram consolidation (Stage~2), per-trajectory fingerprints $\{\mathbf{f}_1, \ldots, \mathbf{f}_N\}$ are z-score normalized per dimension:
$z_k^{(j)} = (f_k^{(j)} - \mu_k) / (\sigma_k + \epsilon)$, where $\mu_k$ and $\sigma_k$ are computed across all $N$ trajectories.
The procedural channel then stores cross-trace statistics---mean, median, standard deviation, min, and max---for each of the 17 features, providing a compact yet informative summary of the profile's behavioral tendencies.

\noindent\textbf{Design rationale.}
The 17 features are chosen to cover all six behavioral dimensions with at least two features each, use only deterministic counting-based computations that require no LLM calls and produce perfectly reproducible outputs, and remain interpretable with a clear behavioral reading per feature. We experimented with higher-dimensional feature sets of up to 50 raw statistics and found that the 17-feature subset retains discriminative power while enabling efficient z-score normalization and deviation detection.


\subsection{Semantic Channel Details}
\label{sec:appendix_semantic}

The semantic channel captures \emph{what} the user produces and \emph{how} they produce it, complementing the procedural channel's quantitative statistics with content-level understanding.

\smallskip\noindent\textbf{Per-trajectory extraction.}
Each Engram's Semantic Unit stores \emph{file metadata}---detected language, file type distribution, naming conventions, and representative filenames---alongside a \emph{behavioral descriptor} generated by a VLM from multimodal file snapshots and edit diffs, summarizing the user's style, formatting, and detail level.
Created-file content and edit-chain diffs are split into 800-character chunks, embedded via Cohere \texttt{embed-english-v3.0} at 1024-D; up to 50 chunks per profile are retained, prioritizing non-deviant trajectories.

\smallskip\noindent\textbf{Cross-session consolidation.}
Stage~2 merges all Semantic Units into a unified profile: aggregated language and naming statistics form the \emph{static content}; an LLM cross-session summary produces unified \emph{Semantic Clues} such as ``produces verbose Markdown reports with structured headers and inline tables''; the embedded chunks are indexed for query-adaptive retrieval at Stage~3.


\subsection{Episodic Segmentation and Boundary Detection}
\label{sec:appendix_segmentation}

The episodic channel partitions each trajectory into 2--5 semantically coherent \emph{episodes}---e.g., ``document survey'' followed by ``report drafting''---and clusters them across trajectories to surface recurrent themes.

\smallskip\noindent\textbf{Per-trajectory segmentation.}
Two LLM calls per trajectory. First, \emph{boundary detection}: the event timeline is rendered as a compact string of ${\sim}$50 chars per event, and an LLM identifies 2--5 focus-shift boundaries, validated, deduplicated, and capped at 4. Trajectories with fewer than 3 events or invalid outputs fall back to a single episode; segments with fewer than 3 events merge with the preceding one. Second, \emph{episode summarization}: for each segment, the LLM generates a title, a third-person narrative of 3--8 sentences, and a one-sentence summary.

\smallskip\noindent\textbf{Cross-trajectory clustering.}
During consolidation, episode summaries are embedded with Cohere \texttt{embed-english-v3.0} at 1024-D and grouped via agglomerative clustering with average linkage, cosine similarity, and a threshold of 0.6, surfacing recurrent themes across sessions.
Separately, trajectories are clustered by their 17-D fingerprints using Euclidean distance with at most 3 clusters to capture distinct behavioral modes such as read-heavy vs.\ production-heavy sessions.
We chose LLM-based segmentation over sliding-window phase detection, which is too coarse to distinguish different episodes within the same phase, and over HMMs, which require labeled transition data unavailable for our task set; non-determinism is mitigated by strict validation and single-episode fallback.


\subsection{Query-Adaptive Retrieval Details}
\label{sec:appendix_retrieval}

Given a query $q$, the retriever concatenates three blocks in fixed order: \emph{Procedural Patterns}---the full L/M/R dimension summary and aggregate statistics, always included; \emph{Semantic Content}---static metadata plus the top-5 content chunks by cosine similarity to $q$ via Cohere \texttt{embed-english-v3.0}; and \emph{Episodic Consistency}---behavioral clusters, anomalous sessions, and the top-5 episode narratives by cosine similarity to $q$.
Content previews are truncated to 800 characters and filenames to 40 characters, as determined by the sensitivity study in \cref{sec:appendix_ablation}. The three channels are concatenated as Markdown sections with no cross-channel re-ranking; ablation experiments confirm all three contribute complementary signal.

\section{Benchmark Construction and Data}
\label{sec:appendix_bench}


\subsection{Profile Design and Instantiation}
\label{sec:appendix_derivation}
\label{sec:appendix_attributes}
\label{sec:appendix_profiles}

\smallskip\noindent\textbf{Dimension derivation.}
We derive the six behavioral dimensions from seven OS-agent use cases---Proactive Assistance, Personalized Defaults, Smart Organization, Context Recovery, Behavioral Continuity, Conflict Detection, and Delegation Quality---by asking \emph{what behavioral aspect must the agent understand} for each, then grouping the resulting needs into orthogonal dimensions. \Cref{tab:derivation} shows the complete mapping.

\begin{table*}[t!]
\caption{\textbf{Dimension derivation.} Each row is an OS-agent use case; each column is a behavioral dimension. Cells indicate the specific capability the dimension enables for that use case.}
\label{tab:derivation}
\centering
\renewcommand{\arraystretch}{1.2}
\setlength{\tabcolsep}{4pt}
\resizebox{\textwidth}{!}{%
\begin{tabular}{@{}lllllll@{}}
\toprule
& \textbf{A: Consump.} & \textbf{B: Product.} & \textbf{C: Organiz.} & \textbf{D: Iteration} & \textbf{E: Curation} & \textbf{F: Cross-M.} \\
\midrule
\textbf{UC1: Proactive}   & Predict next read    & Predict format       & Pre-create dirs      & ---                  & Predict cleanup      & Predict chart need \\
\textbf{UC2: Defaults}     & Set read mode        & Set length \& tone   & Set folder depth     & Set edit granularity & ---                  & Set output modality \\
\textbf{UC3: Smart Org.}   & ---                  & ---                  & Maintain hierarchy   & ---                  & Predict retention    & --- \\
\textbf{UC4: Recovery}     & Reconstruct reads    & Reconstruct drafts   & Navigate folders     & Reconstruct edits    & ---                  & --- \\
\textbf{UC5: Continuity}   & Consistent reading   & Consistent style     & Consistent structure & Consistent editing   & Consistent curation  & Consistent modality \\
\textbf{UC6: Conflict}     & Detect read drift    & Detect style change  & Detect reorganiz.    & Detect edit shift    & Detect curation chg. & --- \\
\textbf{UC7: Delegation}   & Read as user         & Write as user        & Organize as user     & Edit as user         & Curate as user       & Use user modalities \\
\bottomrule
\end{tabular}}
\end{table*}

\smallskip\noindent\textbf{Attribute schema.}
\Cref{tab:attribute_schema} lists all 19 attributes: 3 identity and 16 behavioral, organized under dimensions A--F with L/M/R tiers. Profile Reconstruction evaluates all 16 behavioral attributes per profile, yielding $20 \times 16 = 320$ items.

\begin{table}[t!]
\centering
\caption{\textbf{Profile attribute schema.} Three identity attributes define the user; 16 behavioral attributes span dimensions A--F, each discretized into L/M/R tiers. \texttt{version\_strategy} is shared by dimensions C and D. Profile Reconstruction evaluates all 16 behavioral attributes per profile, yielding $20 \times 16 = 320$ scored items.}
\label{tab:attribute_schema}
\setlength{\tabcolsep}{3pt}
\renewcommand{\arraystretch}{1.05}
\resizebox{\linewidth}{!}{
\begin{tabular}{@{}llccc@{}}
\toprule
\textbf{Attribute} & \textbf{Dim.} & \textbf{L} & \textbf{M} & \textbf{R} \\
\midrule
\texttt{name}       & --- & \multicolumn{3}{c}{Free-form display name} \\
\texttt{role}       & --- & \multicolumn{3}{c}{Professional role} \\
\texttt{language}   & --- & \multicolumn{3}{c}{Primary output language} \\
\midrule
\texttt{reading\_strategy}  & A   & Sequential deep    & Search-first      & Breadth-first \\
\texttt{thoroughness}       & A   & Exhaustive         & Selective          & Minimal \\
\midrule
\texttt{tone}               & B   & Formal, academic   & Professional       & Casual \\
\texttt{output\_detail}     & B   & Comprehensive      & Balanced           & Concise \\
\texttt{output\_structure}  & B   & Highly structured   & Moderate           & Free-form \\
\texttt{documentation}      & B   & Extensive          & Moderate           & Minimal \\
\midrule
\texttt{directory\_style}   & C   & Nested, 3+ levels  & Adaptive, 1--2     & Flat, root only \\
\texttt{naming}             & C   & Systematic         & Semi-structured    & Ad-hoc \\
\texttt{version\_strategy}  & C,D & Explicit v1/v2     & Backup copies      & In-place overwrite \\
\midrule
\texttt{edit\_strategy}     & D   & Incremental edits  & Balanced           & Bulk rewrite \\
\texttt{error\_handling}    & D   & Cautious, backup   & Selective backup   & Direct, no backup \\
\texttt{revision\_depth}    & D   & Multi-pass         & Two-pass           & Single-pass \\
\midrule
\texttt{working\_style}     & E   & Phased, methodical & Pragmatic          & Burst-mode \\
\texttt{cleanup\_policy}    & E   & Aggressive cleanup & Periodic archival  & Never delete \\
\midrule
\texttt{cross\_modal}       & F   & Visual-heavy       & Balanced           & Text-only \\
\texttt{output\_modality}   & F   & Multi-format       & Dual-format        & Single-format \\
\bottomrule
\end{tabular}
}
\end{table}

\smallskip\noindent\textbf{Profile instances.}
\Cref{tab:profile_instances} presents L/M/R assignments for all 20 profiles. Each tier appears in at least 5 profiles per dimension, preventing evaluation bias.

\begin{table*}[t!]
\centering
\caption{\textbf{Profile instances.} L/M/R tier assignments across dimensions A--F for all 20 profiles. Each tier appears in at least 5 profiles per dimension to prevent evaluation bias.}
\label{tab:profile_instances}
\renewcommand{\arraystretch}{1.05}
\setlength{\tabcolsep}{4pt}
\resizebox{\textwidth}{!}{%
\begin{tabular}{@{}cllcccccc|cllcccccc@{}}
\toprule
\textbf{ID} & \textbf{Name} & \textbf{Role} & \textbf{A} & \textbf{B} & \textbf{C} & \textbf{D} & \textbf{E} & \textbf{F} &
\textbf{ID} & \textbf{Name} & \textbf{Role} & \textbf{A} & \textbf{B} & \textbf{C} & \textbf{D} & \textbf{E} & \textbf{F} \\
\midrule
p1  & Chen Wei       & Research Analyst    & L & L & L & L & L & M &
p11 & Priya Sharma   & Supply Chain Ana.   & M & L & L & L & L & R \\
p2  & Liu Jing       & Policy Analyst      & L & L & R & R & L & M &
p12 & Wang Fang      & Journalism Editor   & R & L & R & L & M & M \\
p3  & Sam Taylor     & Ops Manager         & M & R & R & R & M & R &
p13 & Zhao Ming      & Landscape Arch.     & L & M & L & M & L & L \\
p4  & Nakamura Yuki  & Finance Consultant  & M & L & M & L & R & L &
p14 & Daniel Osei    & Compliance Officer  & M & R & L & L & M & R \\
p5  & Maria Santos   & Marketing Coord.    & R & M & M & M & R & M &
p15 & Sophie Laurent & Project Manager     & R & L & M & M & R & M \\
p6  & Alex Kim       & Event Planner       & R & M & L & R & M & R &
p16 & Marcus Chen    & Data Analyst        & M & M & R & M & L & R \\
p7  & Zhang Meilin   & Curriculum Designer & L & M & M & M & M & L &
p17 & Chen Wenjing   & Museum Curator      & L & L & L & L & M & L \\
p8  & Jordan Rivera  & Technical Writer    & R & R & M & L & R & R &
p18 & Aisha Johnson  & Executive Assistant & R & R & R & R & L & M \\
p9  & Li Hao         & UX Researcher       & M & M & L & M & L & L &
p19 & Lin Xiaoyu     & Social Media Mgr.   & M & M & R & M & M & R \\
p10 & Emily Okafor   & Quality Auditor     & L & R & R & R & R & M &
p20 & Tom O'Brien    & Building Inspector  & L & R & M & R & R & L \\
\bottomrule
\end{tabular}}
\end{table*}


\subsection{Task Pool and File Type Statistics}
\label{sec:appendix_tasks}
\label{sec:appendix_filetypes}

We design 32 tasks spanning 6 types---16 text-centric and 16 multimodal with audio, image, or video inputs.
\Cref{tab:task_overview} provides the full task pool.

\smallskip\noindent\textbf{Task representativeness.}
Our six task types---Understand, Organize, Create, Synthesize, Iterate, Maintain---subsume the core desktop activities in OSWorld~\cite{xie2024osworld}, OfficeBench~\cite{wang2024officebench}, and OS-Copilot~\cite{oscopilot}, while adding curation and cross-modal dimensions absent from existing benchmarks.
Code development, real-time collaboration, and system administration are not covered, which we note as a limitation in \cref{sec:appendix_ethics}.

\begin{table*}[ht!]
\centering
\caption{\textbf{Task pool overview.} 32 tasks across 6 types with their activated dimensions and input file composition. A dimension is listed when ${\geq}70\%$ of profiles show non-trivial signal; parenthesized dimensions indicate 30--69\% partial activation. MM marks multimodal tasks.}
\label{tab:task_overview}
\resizebox{\linewidth}{!}{
\begin{tabular}{llcp{7cm}crl}
\toprule
\textbf{Task} & \textbf{Type} & \textbf{MM} & \textbf{Description} & \textbf{Dims} & \textbf{In} & \textbf{Input File Types} \\
\midrule
T-01 & Understand & \xmark & Investment analyst work overview summary & A, B, E, F & 26 & .md(7), .pdf(6), .eml(5), .txt(5), .png(2), .csv(1) \\
T-02 & Understand & \cmark & Legal case materials review and timeline & A, B, E, F & 24 & .eml(6), .pdf(5), .docx(5), .txt(3), .png(2), .xlsx(1), .ics(1), .mp3(1) \\
T-03 & Create & \xmark & Personal knowledge base creation & B, (C), (E) & 0 & \textit{empty workspace} \\
T-04 & Create & \xmark & Meeting minutes and follow-up document creation & B, (E) & 0 & \textit{empty workspace} \\
T-05 & Organize & \xmark & Messy folder cleanup and reorganization & A, B, C, E, F & 30 & .eml(8), .png(4), .md(4), .txt(4), .pdf(4), .jpg(2), .ics(2), .csv(1), .docx(1) \\
T-06 & Synthesize & \xmark & Multi-source synthesis research report & A, B, E, F & 21 & .pdf(8), .eml(6), .txt(3), .md(3), .docx(1) \\
T-07 & Synthesize & \cmark & Diary and notes synthesis into personal profile & A, B, E, F & 22 & .eml(5), .txt(5), .png(4), .mp3(3), .ics(2), .xlsx(1), .pdf(1), .docx(1) \\
T-08 & Create & \xmark & Quarterly work summary report creation & A, B, E, F & 18 & .md(7), .eml(6), .txt(3), .docx(1), .csv(1) \\
T-09 & Iterate & \xmark & Report revision and condensation & B, D, (A), (E) & 1 & .md(1) \\
T-10 & Maintain & \xmark & Knowledge base content update and maintenance & A, B, D, E, F, (C) & 5 & .md(5) \\
T-11 & Iterate & \xmark & Multi-file error detection and correction & A, B, D, E, F & 7 & .md(7) \\
T-12 & Iterate & \xmark & Document format standardization & A, B, D, E, F & 16 & .txt(7), .md(5), .pdf(2), .png(1), .eml(1) \\
T-13 & Iterate & \xmark & Review feedback integration and revision & A, B, D, E, F, (C) & 4 & .md(4) \\
T-14 & Organize & \xmark & Version management and archiving & A, B, C, E, F, (D) & 10 & .md(9), .csv(1) \\
T-15 & Synthesize & \xmark & Conflicting reports analysis and reconciliation & A, B, E, F & 5 & .md(3), .csv(2) \\
T-16 & Understand & \cmark & Time-constrained priority triage & A, E, B, F & 22 & .md(9), .eml(4), .pdf(3), .mp3(2), .png(2), .csv(1), .txt(1) \\
T-17 & Understand & \xmark & File system health check and diagnostics & A, B, E, F & 24 & .eml(6), .pdf(5), .docx(5), .txt(3), .png(2), .xlsx(1), .ics(1), .mp3(1) \\
T-18 & Maintain & \xmark & Legal knowledge base three-round incremental update & A, B, C, D, E, F & 16 & .md(8), .pdf(4), .mp3(1), .docx(1), .eml(1), .png(1) \\
T-19 & Iterate & \cmark & Document audience adaptation & A, B, D, E, F & 16 & .docx(5), .pdf(3), .mp3(3), .eml(3), .md(2) \\
T-20 & Create & \xmark & Weekly report management system setup & B, E, (C) & 0 & \textit{empty workspace} \\
T-21 & Organize & \cmark & File system cleanup and deduplication & C, A, D & 30 & .png(18), .mp3(5), .jpg(5), .tmp(1), .bak(1) \\
T-22 & Understand & \cmark & Film collection catalog and review & A, F, B & 24 & .mp4(13), .gif(6), .jpg(2), .pptx(1), .pdf(1), .docx(1) \\
T-23 & Organize & \cmark & Travel photo album organization & C, F, A & 40 & .jpg(17), .jpeg(16), .png(7) \\
T-24 & Synthesize & \cmark & Earnings call cross-modal analysis & A, B, F & 19 & .mp3(8), .pdf(8), .md(3) \\
T-25 & Understand & \cmark & Legal multimedia evidence review & A, F, B & 25 & .mp4(5), .docx(5), .png(4), .pdf(4), .mp3(4), .eml(3) \\
T-26 & Organize & \cmark & Personal digital asset archiving & C, A, F & 35 & .png(12), .jpg(6), .mp3(5), .mp4(4), .txt(2), .mkv(2), .eml(2), .md(1), .csv(1) \\
T-27 & Create & \cmark & Student portfolio compilation & B, F, C & 25 & .pdf(13), .png(6), .mp4(2), .eml(2), .jpeg(1), .ics(1) \\
T-28 & Synthesize & \cmark & Pet care archive synthesis & A, B, F & 18 & .png(7), .eml(6), .mp3(3), .pdf(1), .ics(1) \\
T-29 & Organize & \cmark & Company registration PDF database & C, D, A & 30 & .pdf(28), .xlsx(2) \\
T-30 & Iterate & \cmark & Voice memo organization and archiving & D, F, A & 17 & .mp3(13), .txt(2), .md(2) \\
T-31 & Create & \cmark & Nature scenery video collection curation & B, F, C & 24 & .mp4(14), .jpeg(9), .jpg(1) \\
T-32 & Maintain & \cmark & Cross-modal archive consistency check & D, A, F & 24 & .png(8), .mp3(4), .mp4(3), .eml(3), .docx(3), .pdf(2), .txt(1) \\
\midrule
\multicolumn{5}{r}{\textbf{Total}} & \textbf{578} & \\
\bottomrule
\end{tabular}
}
\end{table*}


\subsection{Evaluation Pipeline}
\label{sec:appendix_qa}
\label{sec:appendix_cross_validation}

\smallskip\noindent\textbf{QA generation.}
For MCQ tracks, all distractors must share at least 3 dimensions with the target to ensure non-trivial difficulty; GPT-4.1 converts structured templates into natural-language phrasing. For open-ended Profile Reconstruction, an LLM judge scores each attribute on a 1--5 Likert scale---from incorrect identification~\textbf{1} to correct tier with specific evidence~\textbf{5}---with randomized attribute order and calibration examples.

\smallskip\noindent\textbf{Cross-backbone trace validation.}
To verify that behavioral signal is genuine rather than a generation-model artifact, we feed the same \filegramos{} memory context to three QA backbones while fixing the judge to Gemini~2.5-Flash. As shown in \cref{tab:cross_backbone}, all backbones achieve $>$80\% accuracy with $<$2.0\,pp variance, confirming the signal is model-agnostic.

\begin{table}[t!]
\centering
\caption{\textbf{Cross-backbone trace validation.} Per-attribute reconstruction accuracy~(\%) across three QA backbones, all receiving the same \filegramos{} memory context. Inter-backbone variance stays below 2.0\,pp, confirming model-agnostic signal.}
\label{tab:cross_backbone}
\small
\begin{tabular}{@{}lccc@{}}
\toprule
\textbf{QA Backbone} & \textbf{Proc.} & \textbf{Sem.} & \textbf{Avg.} \\
\midrule
Gemini 2.5-Flash  & 84.4 & 80.0 & 82.8 \\
GPT-4.1           & 82.5 & 78.3 & 80.9 \\
Claude Sonnet 4   & 83.8 & 80.0 & 82.2 \\
\bottomrule
\end{tabular}
\end{table}

\section{Extended Experiments and Analysis}
\label{sec:appendix_experiments}


\subsection{Baseline Implementation Details}
\label{sec:appendix_baselines}

All 12 baselines share the same QA backbone---Gemini 2.5-Flash---and receive identical cleaned event logs and output files per profile. \Cref{tab:baseline_details} summarizes each method's category, memory mechanism, and key configuration. All systems use their published default settings; no per-baseline hyperparameter sweeps are performed. For systems not originally designed for file-system traces, such as Mem0 and Zep, each trajectory is mapped to a conversation turn containing the full event log.

\begin{table}[ht!]
\centering
\caption{\textbf{Baseline implementation summary.} All 12 baselines plus \filegramos{} share Gemini 2.5-Flash as QA backbone and receive identical cleaned event logs and output files per profile.}
\label{tab:baseline_details}
\setlength{\tabcolsep}{3pt}
\renewcommand{\arraystretch}{1.05}
\resizebox{\linewidth}{!}{
\begin{tabular}{@{}llll@{}}
\toprule
\textbf{Method} & \textbf{Category} & \textbf{Memory Mechanism} & \textbf{Key Config} \\
\midrule
No Context       & Context  & None                              & Lower bound \\
Full Context     & Context  & Full concatenation                & 625.2K tok avg \\
Naive RAG        & Context  & Chunk embed + top-5 retrieval     & 512-tok chunks, overlap 64 \\
VisRAG           & Context  & ColPali vision embed + top-5      & Page images + text fallback \\
\midrule
Eager Summ.      & Text     & Per-trajectory LLM summary        & Concatenated summaries \\
Mem0             & Text     & Flat key--value store             & Official SDK defaults \\
Zep              & Text     & Graph-based knowledge graph       & Graph + semantic search \\
MemOS            & Text     & Hierarchical tier pipeline        & Working/short/long-term \\
SimpleMem        & Text     & Compact keyword + semantic        & 9.3K tok avg \\
EverMemOS        & Text     & Temporal consolidation + hierarchy & 1098.9K tok avg \\
\midrule
MMA              & Multimodal & Confidence-scored retrieval      & Text + visual ingestion \\
MemU             & Multimodal & VLM captioning + dual store      & PDF/image captioning \\
\midrule
\filegramos{}    & Ours     & 3-channel structured extraction   & $\tau{=}1.5$, 109.7K tok avg \\
\bottomrule
\end{tabular}
}
\end{table}



\subsection{Ablation Studies}
\label{sec:appendix_ablation}

\smallskip\noindent\textbf{Memory channel removal.}
We evaluate \filegramos{} with each channel removed in turn, carefully decoupling shared representations to ensure clean isolation. \Cref{tab:ablation_channel} reports the results.

\begin{table}[t!]
\caption{\textbf{Channel ablation.}
First row: absolute accuracy~(\%); ablation rows: per-cell $\Delta$ relative to the full model. The procedural channel contributes the largest overall drop; all three channels carry distinct, complementary signal.}
\label{tab:ablation_channel}
\centering
\footnotesize
\renewcommand{\arraystretch}{1.1}
\setlength{\tabcolsep}{3pt}
\resizebox{\linewidth}{!}{%
\begin{tabular}{@{}l|cc|cc|cc|ccc|c@{}}
\toprule
\multirow{2}{*}{\textbf{Variant}}
& \multicolumn{2}{c|}{\textbf{T1: Underst.}}
& \multicolumn{2}{c|}{\textbf{T2: Reason.}}
& \multicolumn{2}{c|}{\textbf{T3: Detect.}}
& \multicolumn{3}{c|}{\textbf{Channel}}
& \multirow{2}{*}{\textbf{Avg}} \\
\cmidrule(lr){2-3} \cmidrule(lr){4-5} \cmidrule(lr){6-7} \cmidrule(lr){8-10}
& \makecell[c]{\scriptsize Attr\\\scriptsize Rec} & \makecell[c]{\scriptsize Behav\\\scriptsize FP}
& \makecell[c]{\scriptsize Behav\\\scriptsize Inf} & \makecell[c]{\scriptsize Trace\\\scriptsize Dis}
& \makecell[c]{\scriptsize Anom\\\scriptsize Det} & \makecell[c]{\scriptsize Shift\\\scriptsize Ana}
& \scriptsize Proc & \scriptsize Sem & \scriptsize Epi
& \\
\midrule
\filegramos{} & \textbf{50.6} & \textbf{35.2} & \textbf{42.1} & \textbf{80.9} & \textbf{70.2} & \textbf{37.8} & \textbf{60.1} & \textbf{55.0} & \textbf{58.9} & \textbf{59.6} \\
\midrule
$-$Proc. & $-$5.5 & $-$1.1 & $-$2.3 & \cellcolor{red!15}$-$27.8 & $-$3.9 & \cellcolor{red!10}$-$8.3 & \cellcolor{red!15}$-$12.2 & $-$7.3 & $-$6.8 & $-$11.1 \\
$-$Sem.  & \cellcolor{red!10}$-$10.6 & $-$4.2 & $-$6.1 & $-$2.9 & $-$6.7 & $-$7.8 & $-$3.3 & \cellcolor{red!10}$-$8.8 & $-$4.2 & $-$5.5 \\
$-$Epi.  & $-$5.1 & $-$4.2 & $-$5.1 & $-$1.9 & \cellcolor{red!10}$-$6.2 & $-$5.8 & $-$2.3 & $-$2.5 & \cellcolor{red!10}$-$6.1 & $-$4.2 \\
\bottomrule
\end{tabular}}
\end{table}

The procedural channel is the dominant contributor: removing it causes the largest drop of $-$11.1\,pp, with Trace Disentanglement degrading most severely from 80.9 to 53.1. Removing the semantic or episodic channel produces smaller but meaningful drops of $-$5.5\,pp and $-$4.2\,pp respectively, and each channel's removal most strongly degrades its own question type, validating that the three channels capture genuinely distinct behavioral signals.

\smallskip\noindent\textbf{Parameter sensitivity.}
We vary retrieval-time truncation and context presentation parameters; ingest-time variations all produce identical accuracy and are omitted. \Cref{tab:ablation_param} reports the results.

\begin{table}[ht!]
\caption{\textbf{Parameter sensitivity.}
Per-track accuracy~(\%) on Tracks~1--3. $\Delta$: relative to the 300-char default for retriever rows, and to the 800-char optimum for context rows. Ingest-time parameters have zero effect and are omitted.}
\label{tab:ablation_param}
\centering
\footnotesize
\renewcommand{\arraystretch}{1.05}
\setlength{\tabcolsep}{4pt}
\begin{tabular}{@{}lccccr@{}}
\toprule
\textbf{Configuration} & \textbf{T1} & \textbf{T2} & \textbf{T3} & \textbf{Avg} & \textbf{$\Delta$} \\
\midrule
\multicolumn{6}{@{}l}{\emph{Retriever display length}} \\
300 chars (default) & 46.0 & 70.7 & 48.7 & 53.5 & --- \\
500 chars            & 46.7 & 70.0 & 49.3 & 53.8 & $+$0.3 \\
800 chars            & \textbf{48.0} & 71.3 & 49.3 & \textbf{54.5} & $+$\textbf{1.0} \\
1000 chars           & 46.7 & 72.0 & 49.3 & 54.3 & $+$0.8 \\
\midrule
\multicolumn{6}{@{}l}{\emph{Context presentation}} \\
Preview 200$\to$400 chars    & 47.3 & 72.0 & \textbf{50.7} & 55.2 & $+$0.7 \\
Files/task 3$\to$5           & 46.0 & \textbf{72.7} & 49.7 & 54.5 & $\pm$0.0 \\
Files/task 3$\to$2           & 42.7 & 71.3 & 50.0 & 53.5 & $-$1.0 \\
$+$ Edit chain diffs         & 44.0 & 69.3 & 49.7 & 53.2 & $-$1.3 \\
$-$ Content previews         & 43.3 & 71.3 & 47.3 & 52.3 & $-$2.2 \\
\bottomrule
\end{tabular}
\end{table}

Track~2 is near-invariant across all configurations, as Trace Disentanglement relies on procedural statistics alone. Track~1 is content-sensitive: display at 800 characters yields the best trade-off, while removing content previews degrades it by $-$4.7\,pp. This confirms that procedural features suffice for reasoning, while compact semantic grounding is necessary for attribute inference and change-point detection.

\section{Discussion and Resources}
\label{sec:appendix_discussion}

\subsection{Deployment and System Integration}
\label{sec:appendix_deployment}

Although evaluated on synthetic traces, \filegramos{} is designed for deployment atop real OS-level file-system monitors.

\smallskip\noindent\textbf{Event collection.}
Native APIs---\texttt{FSEvents} on macOS, \texttt{inotify}/\texttt{fanotify} on Linux, \texttt{ReadDirectoryChangesW} on Windows---report file creation, modification, deletion, and renaming in real time with negligible overhead. \filegramos{}'s 12 event types map directly to these notifications. Read-related events such as \texttt{file\_read} and \texttt{file\_browse} additionally require application-level hooks or access-time tracking.

\smallskip\noindent\textbf{Integration architecture.}
A production deployment chains three components: a lightweight \emph{event collector} daemon that filters OS events into a local append-only log; a periodic \emph{Engram encoder} that runs Stage~1 extraction; and an on-demand \emph{memory consolidator + retriever} that updates the three-channel store and assembles query-relevant context. All processing is local by default.

\smallskip\noindent\textbf{Current limitations.}
Key open challenges include interleaved multi-application event streams, duplicate or out-of-order events from cloud-synced file systems, and per-directory privacy opt-in/opt-out controls.

\subsection{Ethical Considerations}
\label{sec:appendix_ethics}

\smallskip\noindent\textbf{Synthetic data and bias.}
All traces are generated by Claude Haiku 4.5 rather than collected from real users, eliminating direct privacy concerns but introducing potential model-inherent biases. We mitigate this through 20 profiles spanning diverse roles, languages, and behavioral configurations, validated by human verifiers. Nonetheless, synthetic traces cannot capture the full complexity of real-world file-system interaction.

\smallskip\noindent\textbf{Privacy.}
File-system traces reveal sensitive patterns---working hours, task priorities, organizational habits---even when synthetically generated. Real-world deployment requires informed consent, data minimization, right to deletion, and access control. \filegramos{} partially addresses minimization by design: the procedural channel stores only 17-D aggregate fingerprints and the semantic channel stores descriptors rather than file contents; however, the episodic channel retains temporal patterns that could be re-identified.

\smallskip\noindent\textbf{Limitations.}
All trajectories originate from a single LLM, which may impose stylistic uniformity absent in real multi-user settings. Behavioral shifts are single-tier perturbations, whereas real drift is often gradual and multi-dimensional. The 32 tasks exclude code development, real-time collaboration, and system administration. With 20 profiles and 640 trajectories the benchmark operates at moderate scale; the sharp accuracy drop in the Real-World setting confirms that sim-to-real transfer remains an open challenge.

\smallskip\noindent\textbf{Intended use.}
\filegrambench{} is a research benchmark for memory and personalization systems, released under a research-use license. It is not intended for surveillance, employee monitoring, or profiling individuals without explicit consent.

\end{document}